\documentclass[draftcls,12pt,onecolumn]{IEEEtran} 

\usepackage{amsfonts}
\usepackage{amsmath}
\usepackage[numbers,sort&compress]{natbib}
\usepackage[colorlinks, linkcolor=red, anchorcolor=blue, citecolor=green]{hyperref}
\usepackage{bm}
\usepackage{graphicx}
\usepackage{subfigure}

\newcommand{\devis}[2]{\textcolor[rgb]{0,0,1}{#2}}
\ifCLASSINFOpdf

\else

\fi

\begin{document}
\title{Deep Learning in Remote Sensing: A Review}

\author{Xiao Xiang Zhu, Devis Tuia, Lichao Mou, Gui-Song Xia, Liangpei Zhang, Feng Xu, Friedrich Fraundorfer

\thanks{X. Zhu and L. Mou are with the Remote Sensing Technology Institute (IMF), German Aerospace Center (DLR), Germany and with Signal Processing in Earth Observation (SiPEO), Technical University of Munich (TUM), Germany, E-mails: xiao.zhu@dlr.de; lichao.mou@dlr.de.}
\thanks{D. Tuia was with  the Department of Geography, University of Zurich, Switzerland. He is now with the Laboratory of GeoInformation Science and Remote Sensing, Wageningen University of Research, the Netherlands. E-mail:  devis.tuia@wur.nl.}
\thanks{G.-S Xia and L. Zhang are with the State Key Laboratory of Information Engineering in Surveying, Mapping and Remote Sensing (LIESMARS), Wuhan University. E-mail:guisong.xia@whu.edu.cn; zlp62@whu.edu.cn.}
\thanks{F. Xu is with the Key Laboratory for Information Science of Electromagnetic Waves (MoE), Fudan Univeristy. E-mail: fengxu@fudan.edu.cn.}
\thanks{F. Fraundorfer is with the Institute of Computer Graphics and Vision, TU Graz, Austria and with the Remote Sensing Technology Institute (IMF), German Aerospace Center (DLR), Germany. E-mail: fraundorfer@icg.tugraz.at.}
\thanks{The work of X. Zhu and L. Mou are supported by the European Research Council (ERC) under the European Union¡¯s Horizon 2020 research and innovation programme (grant agreement No [ERC-2016-StG-714087], Acronym: \textit{So2Sat}), Helmholtz Association under the framework of the Young Investigators Group ``SiPEO'' (VH-NG-1018, www.sipeo.bgu.tum.de) and China Scholarship Council. {The work of D. Tuia is supported by the Swiss National Science Foundation (SNSF) under the project NO. PP0P2 150593.} The work of G.-S. Xia and L. Zhang are supported by the National Natural Science Foundation of China (NSFC) projects with grant No. 41501462 and No. 41431175. The work of F. Xu are supported by the National Natural Science Foundation of China (NSFC) projects with grant No. 61571134. }
}

\markboth{IEEE Geoscience and Remote Sensing Magazine, in press.} 
{Shell \MakeLowercase{\textit{et al.}}: Bare Demo of IEEEtran.cls for IEEE Journals}

\maketitle

\begin{abstract}
\noindent \textbf{This is the pre-acceptance version, to read the final version please go to IEEE Geoscience and Remote Sensing Magazine on IEEE XPlore}.

Standing at the paradigm shift towards data-intensive science, machine learning techniques are becoming increasingly important. In particular, as a major breakthrough in the field, deep learning has proven as an extremely powerful tool in many fields. Shall we embrace deep learning as the key to all? Or, should we resist a ``black-box'' solution? There are controversial opinions in the remote sensing community. In this article, we analyze the challenges of using deep learning for remote sensing data analysis, review the recent advances, and provide resources to make deep learning in remote sensing ridiculously simple to start with. More importantly, we advocate remote sensing scientists to bring their expertise into deep learning, and use it as an implicit general model to tackle unprecedented large-scale influential challenges, such as climate change and urbanization.

\end{abstract}


\IEEEpeerreviewmaketitle

\section{Motivation}
Deep learning is the fastest-growing trend in big data analysis and has been deemed one of the 10 breakthrough technologies of 2013~\cite{MITReview}. It is characterized by neural networks (NNs) involving usually more than two layers (for this reason, they are called \emph{deep}). As their shallow counterpart, deep neural networks exploit feature representations learned exclusively from data, instead of hand-crafting features that are mostly designed based on domain-specific knowledge. Deep learning research has been extensively pushed by Internet companies, such as Google, Baidu, Microsoft, and Facebook for several image analysis tasks, including image indexing, segmentation, and object detection. Recent advances in the field have proven deep learning a very successful set of tools, sometimes even able to surpass human ability to solve highly computational tasks (see, for instance, the highly mediatized Go match between Google's AlphaGo AI and the World Go Champion Lee Sedol. Motivated by those exciting advances, deep learning is becoming the model of choice in many fields of application. For instance, convolutional neural networks (CNNs) have proven to be good at extracting mid- and high-level abstract features from raw images, by interleaving convolutional and pooling layers, (i.e., spatially shrinking the feature maps layer by layer). Recent studies indicate that the feature representations learned by CNNs are greatly effective in large-scale image recognition~\cite{AlexNet,VGGNet,ResNet}, object detection~\cite{RCNN,YOLO}, and semantic segmentation~\cite{FCN,Deconv4Seg2015}. Furthermore, as an important branch of the deep learning family, recurrent neural networks (RNNs) have been shown to be very successful on a variety of tasks involved in sequential data analysis, such as action recognition~\cite{Donahue2015,Du2015} and image captioning~\cite{KXu}.
\par
Following this wave of success and thanks to the increased availability of data and computational resources, the use of deep learning in remote sensing is finally taking off in remote sensing as well. Remote sensing data bring some new challenges for deep learning, since satellite image analysis raises some unique questions that translate into challenging new scientific questions:

\begin{itemize}
\item Remote sensing data are often \textit{multi-modal}, e.g. from optical (multi- and hyperspectral) and synthetic aperture radar (SAR) sensors, where both the imaging geometries and the content are completely different. Data and information fusion uses these complementary data sources in a synergistic way. Already prior to a joint information extraction, a crucial step is to develop novel architectures for the matching of images taken from different perspectives and even different imaging modality, preferably without requiring an existing 3D model. Also, besides conventional decision fusion, an alternative is to investigate the transferability of trained networks to other imaging modalities. 

\item Remote sensing data are \emph{geo-located}, i.e., they are naturally located in the geographical space. Each pixel corresponds to a spatial coordinate, which facilitates the fusion of pixel information with other sources of data, such as GIS layers, geo-tagged images from social media, or simply other sensors (as above). On one hand, this fact allows tackling of data fusion with non-traditional data modalities while, on the other hand, it opens the field to new applications, such as pictures localization, location-based services or reality augmentation.

\item Remote Sensing data are \textit{geodetic measurements} with controlled quality. This enables us to retrieve geo-parameters  with confidence estimates. However, differently from purely data-driven approaches, the role of prior knowledge about the sensors adequacy and data quality becomes even more crucial. For example, to retrieve topographic information, even at the same spatial resolution, interferograms acquired using single-pass SAR system are considered to be more important than the ones acquired in repeat-pass manner.

\item \textit{The time variable} is becoming increasingly in the field. The Copernicus program guarantees continuous data acquisition for decades. For instances, Sentinel-1 images the entire Earth every six days. This capability is triggering a shift from individual image analysis to time-series processing. Novel network architectures must be developed for optimally exploiting the temporal information jointly with the spatial and spectral information of these data.

\item Remote sensing also faces the \textit{big data challenge}. In the Copernicus era, we are dealing with very large and ever-growing data volumes, and often on a global scale. For example, even if they were launched in 2014, Sentinel satellites have already acquired about 25 Peta Bytes of data. The Copernicus concept calls for global applications, i.e., algorithms must be fast enough and sufficiently transferrable to be applied for the whole Earth surface. On the other hand, these data are well annotated and contain plenty of metadata. Hence, in some cases, large training data sets might be generated (semi-) automatically.

\item In many cases remote sensing aims at retrieving \textit{geo-physical or bio-chemical quantities} rather than detecting or classifying objects. These quantities include mass movement rates, mineral composition of soils, water constituents, atmospheric trace gas concentrations, and terrain elevation of biomass. Often process models and expert knowledge exist that is traditionally used as priors for the estimates. This particularity suggests that the so-far dogma of expert-free fully automated deep learning should be questioned for remote sensing and physical models should be re-introduced into the concept, as, for example, in the concept of emulators~\cite{Riv15}.
\end{itemize}
\par

Remote sensing scientists have exploited the power of deep learning to tackle these different challenges and started a new wave of promising research. In this paper, we review these advances. After the introductory Section~\ref{sec:intro} detailing deep learning models (with emphasis put on convolutional neural networks), we enter sections dedicated to advances in hyperspectral image analysis (Section~\ref{sec:hsi}), synthetic aperture radar (Section~\ref{sec:sar}), very high resolution (Section~\ref{sec:vhr}, data fusion (Section~\ref{sec:df}), and 3D reconstruction (Section~\ref{sec:3D}). Section~\ref{sec:tools} then provides the tools of the trade for scientists willing to explore deep learning in their research, including open codes and data repositories. Section~\ref{sec:concl} concludes the paper by giving an overview of the challenges ahead.

\section{From Perceptron to Deep Learning}\label{sec:intro}
Perceptron is the basic of the earliest NNs~\cite{NN1989}. It is a bio-inspired model for binary classification that aims to mathematically formalize how a biological neuron works. In contrast, deep learning has provided more sophisticated methodologies to train deep NN architectures. In this section, we recall the classic deep learning architectures used in visual data processing.
\par
\subsection{Autoencoder models}

\subsubsection{Autoencoder and Stacked Autoencoder (SAE)}
An autoencoder~\cite{ae} takes an input $\bm{x}\in \mathbb{R}^{D}$ and, first, maps it to a latent representation $\bm{h}\in \mathbb{R}^{M}$ via a nonlinear mapping:
\begin{equation}
\bm{h}=f(\bm{\Theta} \bm{x}+\bm{\beta})\,,
\end{equation}
where $\bm{\Theta}$ is a weight matrix to be estimated during training, $\bm{\beta}$ is a bias vector, and $f$ stands for a nonlinear function, such as the logistic sigmoid function or a hyperbolic tangent function. The encoded feature representation $\bm{h}$ is then used to reconstruct the input $\bm{x}$ by a reverse mapping leading to the reconstructed input $\bm{y}$:
\begin{equation}
\bm{y}=f(\bm{\Theta}' \bm{h}+\bm{\beta}')\,,
\end{equation}
where $\bm{\Theta}'$ is usually constrained to be the form of $\bm{\Theta}'=\bm{\Theta}^{T}$, i.e., the same weight is used for encoding the input and decoding the latent representation. The reconstruction error is defined as the Euclidian distance between $\bm{x}$ and $\bm{y}$ that is constrained to approximate the input data $\bm{x}$ (i.e., making $\|\bm{x}-\bm{y}\|_{2}^{2}\rightarrow 0$). The parameters of the autoencoder are generally optimized by stochastic gradient descent (SGD).
\par
An SAE is a neural network consisting of multiple layers of autoencoders in which the outputs of each layer are wired to the inputs of the following one.

\subsubsection{Sparse Autoencoder} The conventional autoencoder relies on the dimension of the latent representation $\bm{h}$ being smaller than that of input $\bm{x}$, i.e., $M<D$, which means that it tends to learn a low-dimensional, compressed representation. However, when $M>D$, one can still discover interesting structures by enforcing a sparsity constraint on the hidden units. Formally, given a set of unlabeled data $\bm{X} = \{\bm{x}^1,\bm{x}^2,\cdots,\bm{x}^N\}$, training a sparse autoencoder~\cite{sparseAE} boils down to finding the optimal parameters by minimizing the following loss function:
\begin{equation}
\mathbb{E} = \frac{1}{N}\sum_{i=1}^{N}(J(\bm{x}^i,\bm{y}^i;\bm{\Theta},\bm{\beta})+\lambda\sum_{j=1}^M\mathrm{KL}(\rho\|\hat{\rho}_j))\,,
\end{equation}
where $J(\bm{x}^i,\bm{y}^i;\bm{\Theta},\bm{\beta})$ is an average sum-of-squares error term, which represents the reconstruction error between the input $\bm{x}^i$ and its reconstruction $\bm{y}^i$. $\mathrm{KL}(\rho\|\hat{\rho}_j)$ is the Kullback-Leibler (KL) divergence between a Bernoulli random variable with mean $\rho$ and a Bernoulli random variable with mean $\hat{\rho}_j$. KL-divergence is a standard function for measuring how similar two distributions are\devis{, and it can be described as follows:}{:}
\begin{equation}
\mathrm{KL}(\rho\|\hat{\rho}_j)=\rho\log\frac{\rho}{\hat{\rho}_j}+(1-\rho)\log\frac{1-\rho}{1-\hat{\rho}_j}\,.
\end{equation}
\par

In the sparse autoencoder model, the KL-divergence is a sparsity penalty term, and $\lambda$ controls its importance. $\rho$ is a free parameter corresponding to a desired average activation\footnote{{An activation corresponds to how much a region of the image reacts when convolved with a filter. In the first layer, for example, each location in the image receives a value that corresponds to a linear combination of the original bands and the filter applied. The higher such value, the more `activated' this filter is on that region. When convolved over the whole image, a filter produces an activation map, which is the activation at each location where the filter has been applied.}} value, and $\hat{\rho}$ indicates the average activation value of hidden neuron $\bm{h}_j$ over the training samples. Similar to the autoencoder, the optimization of a sparse autoencoder can be achieved via back-propagation and SGD.

\begin{figure}[t]
\centering
\includegraphics[width=0.95\columnwidth]{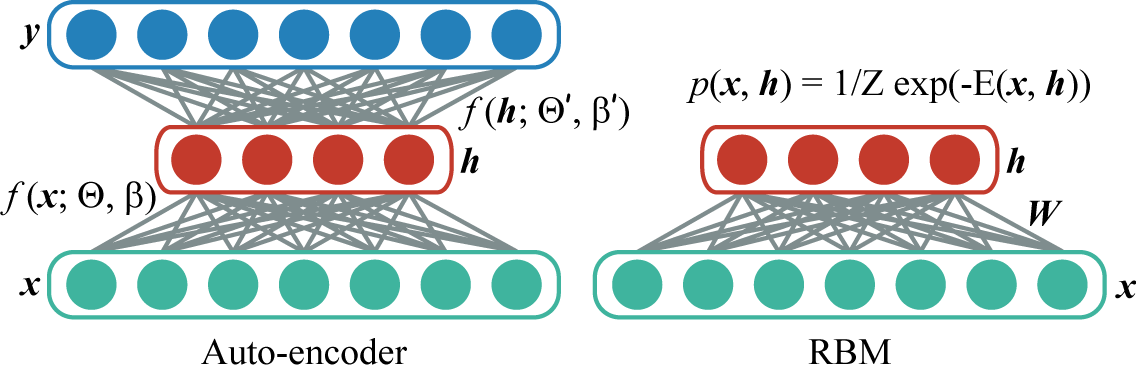}\\
\caption{{Schematic comparison of an autoencoder (left) versus a restricted Boltzmann Machine (right).}} \label{fig:ae_rbm}
\end{figure}

\subsubsection{Restricted Boltzmann Machine (RBM) \& Deep Belief Network (DBN)}
Unlike the deterministic network architectures, such as autoencoders or sparse autoencoders, an RBM {(cf.~Fig.~\ref{fig:ae_rbm})} is a stochastic undirected graphical model consisting of a visible layer and a hidden layer, and it has symmetric connections between these two layers. No connecting exists within the hidden layer or the input layer. The energy function of an RBM can be defined as follows:
\begin{equation}
\mathbb{E}(\bm{x},\bm{h})=\frac{1}{2}\bm{x}^{T}\bm{x}-(\bm{h}^{T}\bm{W}\bm{x}+\bm{c}^{T}\bm{x}+\bm{b}^{T}\bm{h})\,,
\end{equation}
where $\bm{W}$, $\bm{c}$, and $\bm{b}$ are learnable weights\devis{ of the RBM model}{}. Here, the input $\bm{x}$ is also named as the visible random variable, which is denoted as $\bm{v}$ in~\cite{dbn}. The joint probability distribution of the RBM is defined as:
\begin{equation}
p(\bm{x},\bm{h})=\frac{1}{Z}\exp(-\mathbb{E}(\bm{x},\bm{h}))\,,
\end{equation}
where $Z$ is a normalization constant. The form of the RBM makes the conditional probability distribution computationally feasible, when $\bm{x}$ or $\bm{h}$ are fixed.
\par
The feature representation ability of a single RBM is limited. However, its real power emerges when a couple of RBMs are stacked, forming a DBN~\cite{dbn}. Hinton \textit{et al.}~\cite{dbn} proposed a greedy approach that trains RBM in each layer to efficiently train the whole DBN.

\subsection{Convolutional neural networks (CNNs).}\label{sec:cnns}
Unsupervised deep neural networks have been under the spot in the recent year. The leading model is the convolutional neural network (CNN), which learns the filters performing convolutions in the image domain. Here, we briefly review some successful CNN architectures proposed in computer vision in the recent years. For a comprehensive introduction on CNNs, we invite the reader to consider the excellent book by Goodfellow and colleagues~\cite{Goodfellow-et-al-2016}. 

\begin{figure}[t]
\centering
\includegraphics[width=\columnwidth]{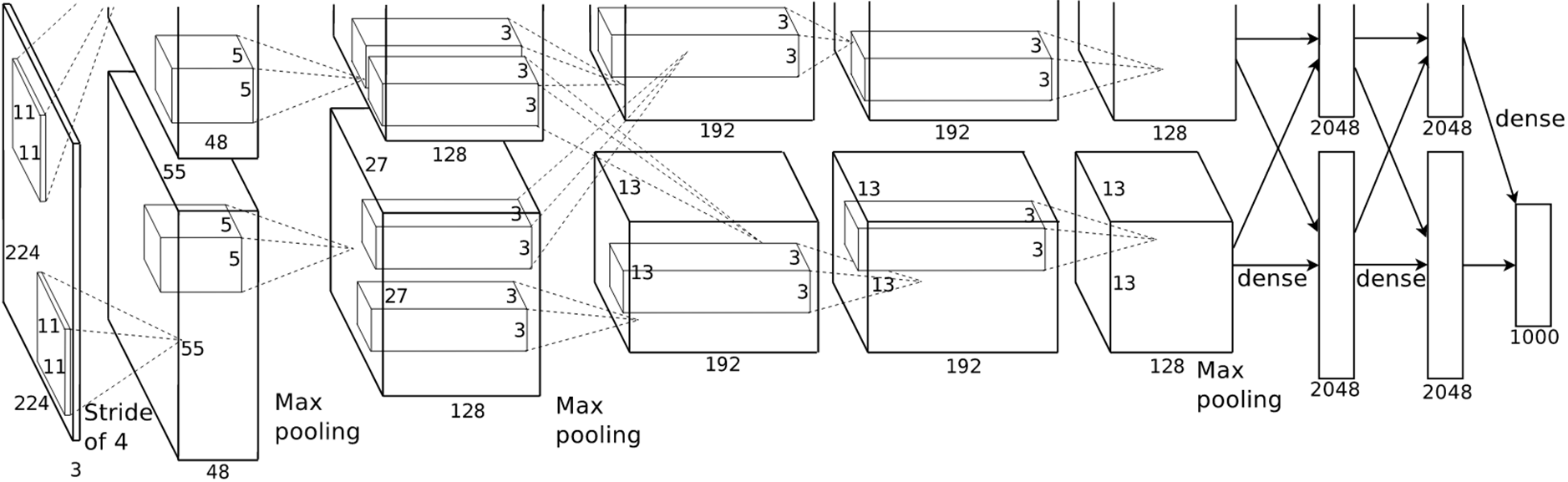}\\
\caption{Architecture of AlexNet, as shown in~\cite{AlexNet}.} \label{fig:alexnet}
\end{figure}

\subsubsection{AlexNet}
In 2012, Krizhevsky \emph{et al.}~\cite{AlexNet} created AlexNet, which is a ``large, deep convolutional neural network'' that won the 2012 ILSVRC (ImageNet Large-Scale Visual Recognition Challenge). The year 2012 is marked as the first year where a CNN was used to achieve a top 5 test error rate of 15.4\%.
\par
AlexNet (cf. Fig.~\ref{fig:alexnet}) scaled the insights of LeNet~\cite{LeNet} into a deeper and much larger network that could be used to learn the appearance of more numerous and more complicated objects. The contributions of AlexNet are as follows:
\begin{itemize}
\item[-] Using rectified linear units (ReLU) as nonlinearity functions that are capable of decreasing training time, as ReLU is several times faster than the conventional hyperbolic tangent function.
  \item[-] Implementing dropout layers in order to avoid the problem of overfitting.
  \item[-] Using data augmentation techniques to artificially increase the size of the training set (and see a  more diverse set of situations). From this, the training patches are translated and reflected on the horizontal and vertical axes.

\end{itemize}

One of the keys of the success of AlexNet is that the model was trained on GPUs. Since GPUs can offer a much larger number of cores than CPUs, it allows much faster training, which in turn allows one to use larger datasets and bigger images. 

\subsubsection{VGG Net}
The design philosophy of the VGG Nets~\cite{VGGNet} is simplicity and depth. In 2014, Simonyan and Zisserman created VGG Nets that strictly makes use of $3\times 3$ filters with stride and padding of 1, along with $2\times 2$ max-pooling layers with stride 2. The main points of VGG Nets are that they:
\begin{itemize}
  \item[-] Use filters with small receptive field of $3\times 3$, rather than using larger ones ($5\times 5$ or $7\times 7$, as in Alexnet).
  \item[-] Have the same feature map size and number of filters in each convolutional layer of the same block.
  \item[-] Increase the size of the features in the deeper layers, roughly doubling after each max-pooling layer.
  \item[-] Use scale jittering as one data augmentation technique during training.
\end{itemize}

VGG is one of the most influential CNN models, as it reinforces the notion that CNNs with deeper architectures can promote hierarchical feature representations of visual data, which in turn improves the classification accuracy. A drawback is that, to train such a model from scratch, one would need large computational power and a very large labeled training set.

\subsubsection{ResNet}

He \emph{et al.}~\cite{ResNet} pushed the idea of very deep networks even further by proposing the 152-layers ResNet -- which won ILSVRC 2015 with an error rate of 3.6\% and set new records in classification, detection, and localization through a single network architecture. In~\cite{ResNet}, authors provide an in-depth analysis about the degradation problem, i.e., simply increasing the number of layers in plain networks results in higher training and test errors, and claim that it is easier to optimize the residual mapping in the ResNet than to optimize the original, unreferenced mapping in the conventional CNNs. The core idea of ResNet is to add shortcut connections that by-pass two or more stacked convolutional layers by performing identity mapping, which are then added together with the output of stacked convolutions.

\begin{figure}[t]
\centering
\includegraphics[width=\columnwidth]{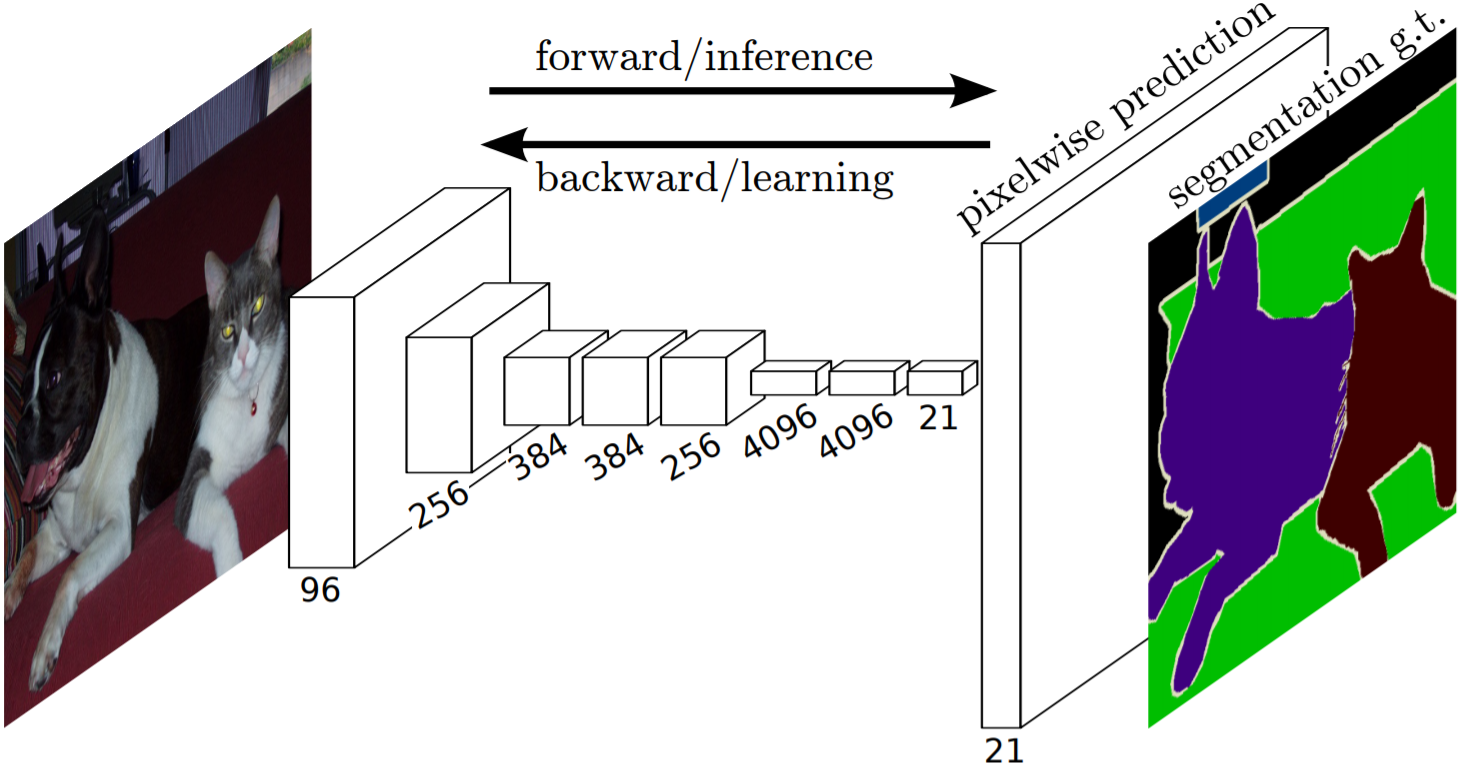}\\
\caption{Architecture of FCN~\cite{FCN}.} \label{fig:fcn}
\end{figure}

\subsubsection{FCN}
The fully convolutional network (FCN)~\cite{FCN} is the most important work in deep learning for semantic segmentation, which is the task of assigning a semantic label to every pixel in the image. To perform this task, the output of the CNN must be of the same pixels size as the input (contrarily to the `single class per image' of the aforementioned models). FCN introduces many significant ideas:
\begin{itemize}
	\item[-] End-to-end learning of the upsampling algorithm via an encoder/decoder structure that first downsamples the activations size and then upsamples it again.
    \item[-] Using fully convolutional architecture allows the network to take images of arbitrary size as input since there is no fully connected layer at the end that requires a specific size of the activations.

    \item[-] Introducing skip connections as a way of fusing information from different depths in the network for the multi-scale inference.
\end{itemize}
\par
Fig.~\ref{fig:fcn} shows the architecture of FCN.

\section{Remote Sensing Meets Deep Learning}

\begin{figure}[t]
\centering
\includegraphics[width=0.7\columnwidth]{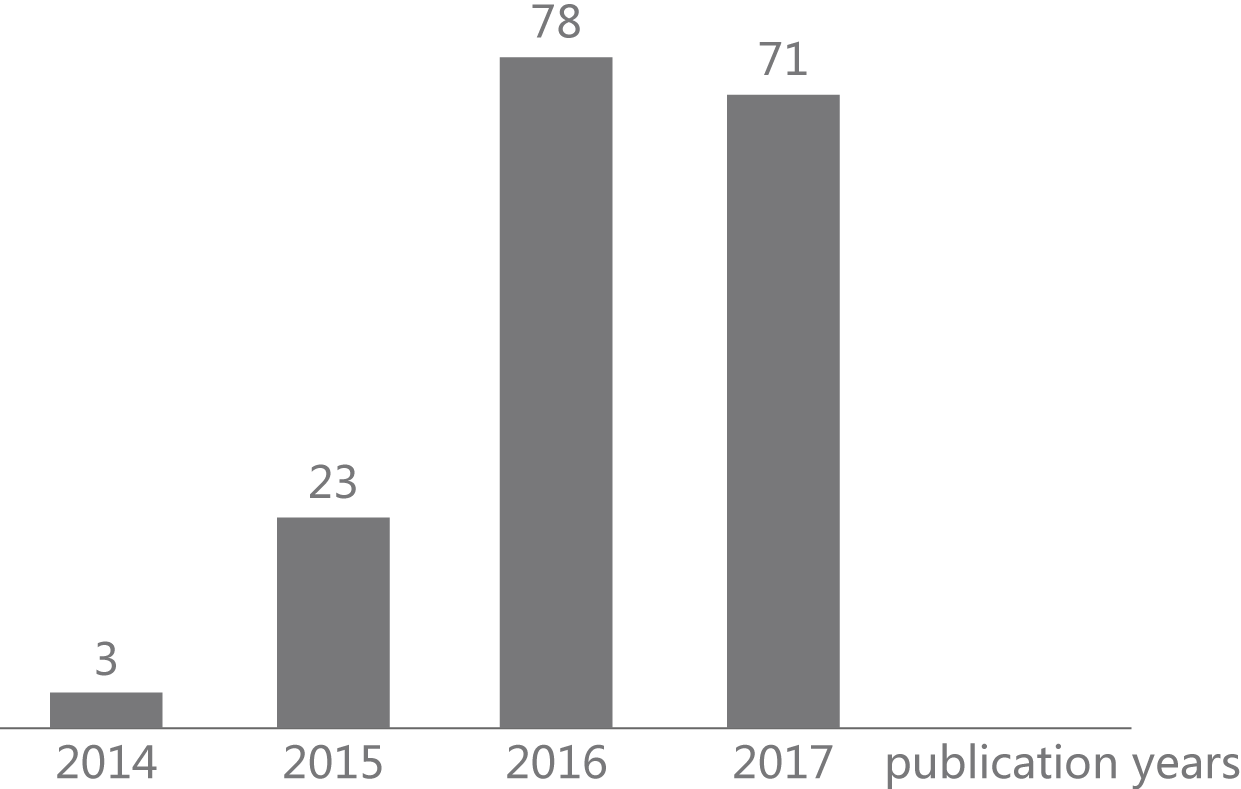}
\caption{Statistics on papers related to deep learning in remote sensing. [source: ISI web of Science; status: September 2017]} \label{fig:isi_figures}
\end{figure}

Deep learning is taking off in remote sensing, as shown in Fig.~\ref{fig:isi_figures}, which summarizes the number of papers on the topic since 2014. Their exponential increase confirms the rapid surge of interest in deep learning for remote sensing. In this section, we focus on a variety of remote sensing applications that are achieved by deep learning and provide an in-depth investigation from the perspectives of hyperspectral image analysis, interpretation of SAR images, interpretation of high-resolution satellite images, multimodal data fusion, and 3D reconstruction.

\subsection{Hyperspectral Image Analysis}\label{sec:hsi}
Hyperspectral sensors are characterized by hundreds of narrow spectral bands. This very high spectral resolution enables us to identify the materials contained in the pixel via spectroscopic analysis. Analysis of hyperspectral data is of high importance in many practical applications, such as land cover/use classification or change and object detection. Also, because high quality hyperspectral satellite data is becoming available, e.g., via the launch of EnMAP, planned in 2020, and DESIS, planned in 2017, hyperspectral image analysis has been one of the most active research directions in the remote sensing community over the last decade.
\par
Inspired by the success of deep learning in computer vision, preliminary studies have been carried out on deep learning in hyperspectral data analysis, which brings new momentum into this field. In this section, we would like to review two application cases, namely, land cover/use classification~(\ref{sec:hsi_classif}) and anomaly detection~(\ref{sec:hsi_ad}).

\begin{figure*}[htb!]
\centering
\includegraphics[width=0.9\linewidth]{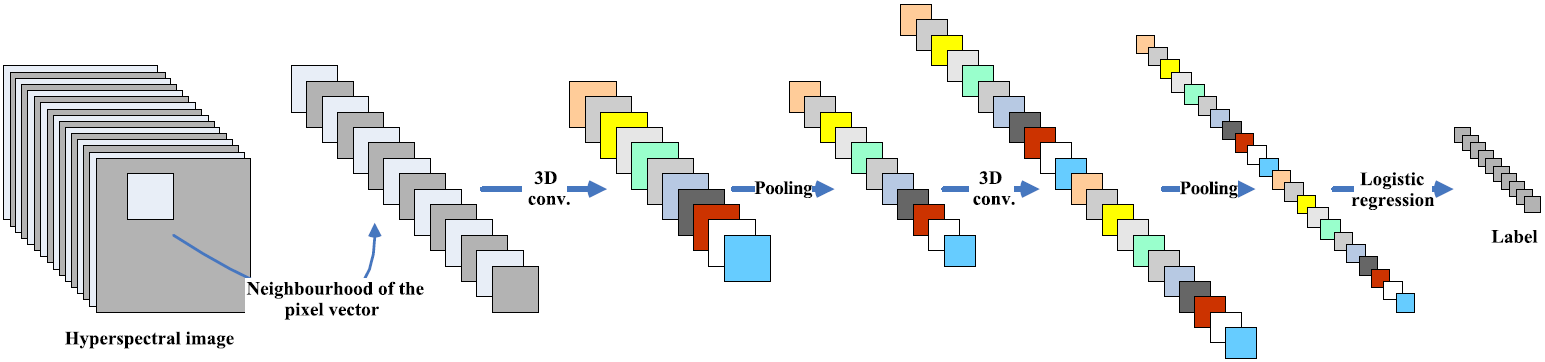}
\caption{Flowchart of the 3D CNN architecture proposed in ~\cite{Yushi16cnn} for spectral-spatial hyperspectral image classification.  \label{fig:3dcnn4classifi}}
\end{figure*}

\subsubsection{Hyperspectral Image Classification}
\label{sec:hsi_classif}
Supervised classification is probably the most active research area in hyperspectral data analysis. There is a vast literature on this topic using the conventional supervised machine learning models, such as decision trees, random forests, and support vector machines (SVMs)~\cite{Cam13}. With the investigation of hyperspectral image classification~\cite{pedram16selfcnn}, a major finding was that various atmospheric scattering conditions, complicated light scattering mechanisms, inter-class similarity, and intra-class variability result in the hyperspectral imaging procedure being inherently nonlinear. It is believed that, in comparison to the aforementioned ``shallow'' models, deep learning architectures are able to extract high-level, hierarchical, and abstract features, which are generally more robust to the nonlinear input data.
\par
 i) Autoencoders for hyperspectral data classification: A first attempt in this direction can be found in~\cite{Yushi14ae}, where authors make use of a stacked autoencoder to extract hierarchical features in the spectral domain. Subsequently, in~\cite{Yushi15dbn}, authors employ DBM. Similarly, Tao \emph{et al.}~\cite{Tao15ssae} use sparse stacked autoencoder to learn an effective feature representation from unlabeled data, and then the learned features are fed into a linear SVM for hyperspectral data classification.
\par
ii) Supervised CNNs: 
In~\cite{WLiSensors}, authors train a simple 1D CNN that contains five layers, namely, an input layer,  a convolutional layer, a max pooling layer, a fully connected layer, and an output layer -- and directly classify the hyperspectral images in spectral domain.
\par
Makantasis~\textit{et al.}~\cite{Makantasis15} exploited a 2D CNN to encode spectral and spatial information, followed by a multi-layer perceptron performing the actual classification.
In~\cite{Kussul17}, authors attempt to carry out classification of crop types using 1D CNN and 2D CNN. They concluded that the 2D CNNs can outperform the 1D CNNs, but some small objects in the final classification map provided by 2D CNN are smoothed and misclassified. 
To avoid overfitting, Zhao and Du~\cite{Zhao16TGRS} propose a spectral-spatial feature-based classification framework, which jointly makes use of a local discriminant embedding-based dimension reduction algorithm and a 2D CNN. In~\cite{pedram16selfcnn}, authors propose a self-improving CNN model, which combines a 2D CNN with a fractional order Darwinian particle swarm optimization algorithm to iteratively select the most informative bands that are suitable for training the designed CNN. {Santara~\textit{et al.}~\cite{Santara17TGRS} propose an end-to-end band-adaptive spectral-spatial feature learning network to address the problems of the curse of dimensionality. In~\cite{Guodong17TGRS}, to allow CNN appropriately trained using limited labeled data, authors present a novel pixel-pair CNN to significantly augment the number of training samples.} 
\par
{Following recent vision developments in 3D CNNs~\cite{C3D}, in which the third dimension usually refers to the time axis, such architecture has also been employed in hyperspectral classification. In other words, in 3D CNN, convolution operations are performed spatial-spectrally while in 2D CNNs they are done only spatially. Compared to 1D and 2D CNNs, 3D CNNs can model spectral information better owing to 3D convolution operations. Authors in~\cite{Yushi16cnn} introduced a supervised, $\ell_2$ regularized 3D CNN-based model 
(see Fig.~\ref{fig:3dcnn4classifi}). While authors of \cite{YingLi16} followed a similar idea for spatial-spectral classification.} 

\begin{figure}[ht!]
\centering
\subfigure[]{
\includegraphics[width=0.43\columnwidth]{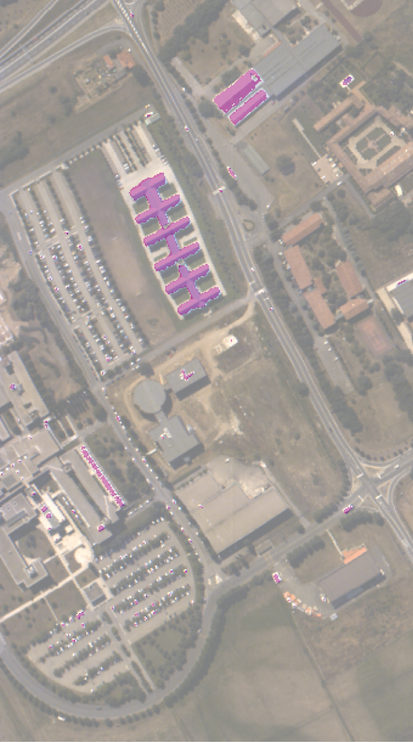}}
\subfigure[]{
\includegraphics[width=0.43\columnwidth]{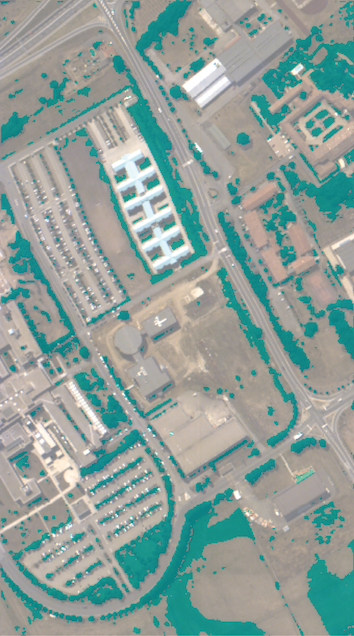}}
\renewcommand{\figurename}{Fig}
\caption{\label{fig:det} Object detection maps using learned filters of the first residual block in the unsupervised Residual Conv-Deconv network~\cite{MouConvDeconv17,MouConvDeconvTGRS17}, in which some ``neurons'' own good description power for semantic visual patterns in the object level. For example, the feature maps activated by the convolutional filters \# 52 and \# 03 in the first residual block can be used to precisely capture (a) metal sheets and (b) vegetative covers, respectively.}
\end{figure}

\par
iii) Unsupervised Deep Learning: To be less dependent on the existence of large annotated collections of labeled data, unsupervised feature extraction remains of great interest. Authors of~\cite{romero16} propose an unsupervised convolutional network for learning spectral-spatial features using sparse learning to estimate the network weights in a greedy layer-wise fashion instead of end-to-end learning. Mou \textit{et al.}~\cite{MouConvDeconv17,MouConvDeconvTGRS17} propose a network architecture called fully Residual Conv-Deconv network for unsupervised spectral-spatial feature learning of hyperspectral images. They report an extensive study of the filters learned (cf. Fig.~\ref{fig:det}).

\begin{figure*}[htb!]
\centering
\includegraphics[width=0.75\linewidth]{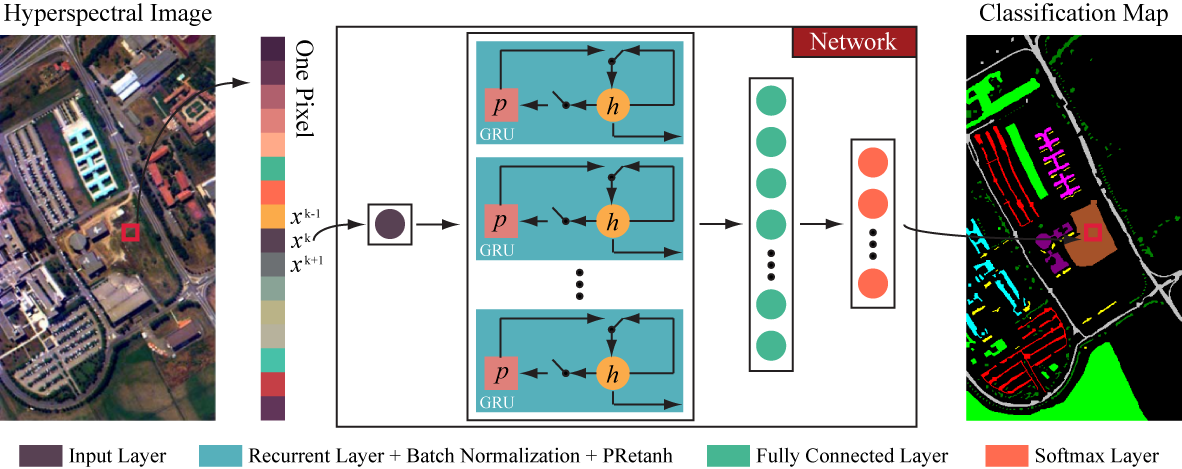}
\caption{RNN proposed for hyperspectral image classification  task in~\cite{Mournn}.\label{fig:rnn4classifi}}
\end{figure*}

\par
iv) RNN for Hyperspectral Image Classification: 
In~\cite{Mournn}, authors propose a  RNN model with a new activation function and modified gated recurrent unit for hyperspectral image classification, which can effectively analyze hyperspectral pixels as sequential data and then determine information categories via network reasoning (cf. Fig.~\ref{fig:rnn4classifi}).

\subsubsection{Anomaly Detection}
\label{sec:hsi_ad}
In a hyperspectral image, the pixels whose spectral signatures are significantly different from the global background pixels are considered anomalies. Since the prior knowledge of the anomalous spectrum is difficult to obtain in practice, anomaly detection is usually solved by background modeling or statistical characterization for hyperspectral data. So far, the only mark addressing this problem via deep learning can be found in ~\cite{WeiLi17}. Li \textit{et al.}~\cite{WeiLi17} propose an anomaly detection framework, in which a multi-layer CNN is trained by using the differences in gray values between neighboring pixel pairs in the reference image as input data. Then, in the test phase, anomalies are detected by evaluating differences between neighboring pixel pairs using the trained CNN.

In summary, deep learning has been widely applied to the multi/hyper-spectral image classification, and some promising results have been achieved. In contrast, for other hyperspectral data analysis tasks, such as change and anomaly detections, deep learning has just made its mark~\cite{Lyu,WeiLi17}. Some potential problems to be further explored include nonlinear spectral unmixing, hyperspectral image enhancement, hyperspectral time-series analysis, etc.

\subsection{Interpretation of SAR Images}\label{sec:sar}

\par
Over the past few years, there have been many publications in deep learning-related studies for SAR image analysis. Among these studies, deep learning techniques have been mostly applied in typical applications, including automatic target recognition (ATR), terrain surface classification, and parameter inversion. This section reviews some of the relevant studies in this area.

\subsubsection{Automatic Target Recognition}
SAR ATR is an important application, in particular for military surveillance~\cite{Dudgeon1993}. A standard architecture for efficient ATR consists of three stages: detection, discrimination, and classification. Each stage tends to perform a much complicated and refined processing than its predecessor, and selects the candidate objects for the next stage processing. However, all three stages can be treated as a  classification problem and, for this reason, deep learning has found its marks. \par

\par
Chen \emph{et al.}~\cite{ChenATR2014} introduce CNN into SAR ATR and tested on the standard ATR dataset MSTAR~\cite{Keydel1996}. The major issue is found to be the lack of sufficient training samples as compared to optical images. It might cause severe overfitting and, therefore, greatly limits the capability of generalizing the model. Data augmentation is employed {to counteract overfitting. Chen \emph{et al.}~\cite{ChenTGRS2016} propose to further remove all fully-connected layers from conventional CNNs which are accountable for most trainable parameters. The final performance is demonstrated superior on the MSTAR dataset (i.e., a state-of-the-art accuracy of 99.1\% on standard operating condition (SOC)). Extensive experiments are conducted to test the generalization capability of the so-called AConvNets and they are found to be quite robust in  several extended operating conditions (EOC). The removal of the fully-connected layers, which is originally designed to be a trainable classifier, might be justifiable in this case, because the limited number of target types can be seen as the feature templates that the AConvNets is extracting.
\par
Many authors applied CNN to SAR ATR and tested on MSTAR dataset, e.g., \cite{Morgan2015,DingGRSL2016,DuRSL2016,Wilmanski2016}, etc. Among these studies, the one common finding is that data augmentation is necessary and the most critical step for SAR ATR using CNNs. Various augmentation strategies are proposed, including translation, rotation, interpolation, etc.
\par
Cui \emph{et al.}~\cite{CuiMPE2016} introduce DBN to SAR ATR, where stacked RBM are used to extract features and then fed to trainable classifier.
\par
Wagner~\cite{WagnerTGRS2016} proposes to use CNN to first extract feature vectors and then fed them to a SVM for classification. The CNN is trained with a fully-connected layer but only the previous activations are used. A systematic data augmentation approach is employed, which includes elastic distortions and affine transformations. It is intended to mimic  typical imaging errors, such as a changing range (which is scale dependent on the depression angle) or an incorrectly estimated aspect angle.
\par
More studies applying CNNs to the ART problem are found. Bentes \emph{et al.}~\cite{BentesEUSAR2016} apply CNN to ship-iceberg discrimination and tested on TerraSAR-X StripMap images. Schwegmann \emph{et al.}~\cite{Schwegmann} apply a specific type of deep neural networks, the highway networks, to ship discrimination in SAR imagery and achieved promising results.
{\O}degaard \emph{et al.}~\cite{Odegaard2016} apply CNN to detect ships in harbor background in SAR images. To address the issue of lack of training samples, they employed a simulation software to generate simulated data for training. 
 Song \emph{et al.}~\cite{SongIGARSS2017} follow this idea and introduce a deep generative neural network for SAR ATR. A generative deconvolutional NN is first trained to generate simulated SAR image from a given target label, during which a feature space is constructed in the intermediate layer. A CNN is then trained to map an input SAR image to the feature space. The goal is to develop an extended ATR system which is capable of interpreting a previously unseen target in the context of all known targets.

\begin{figure*}[tbh!]
\centering
\includegraphics[width=0.9\linewidth]{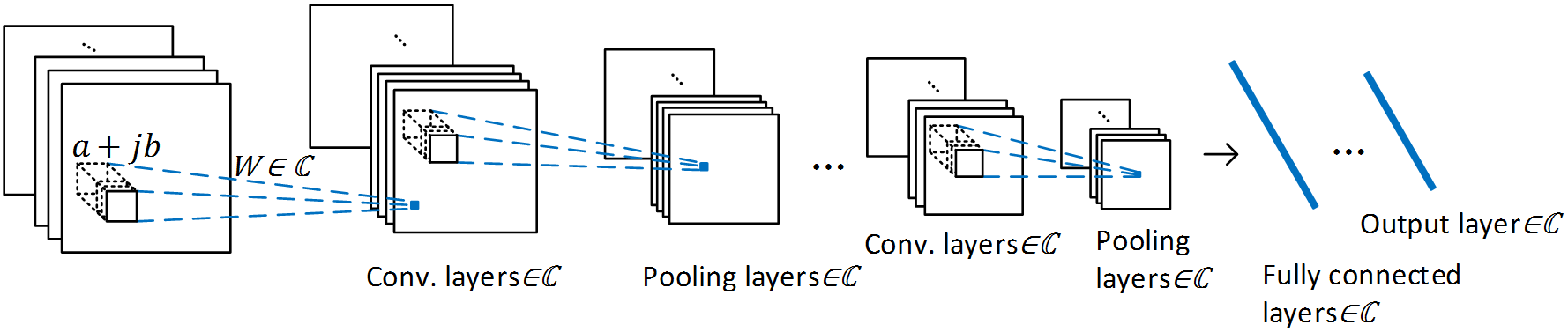}
\caption{Structure of complex-valued CNN (adapted from~\cite{ZZhang2017}).}
\label{fig:cv_cnn}
\end{figure*}

\begin{figure}[tbh!]
\centering
\includegraphics[width=\columnwidth]{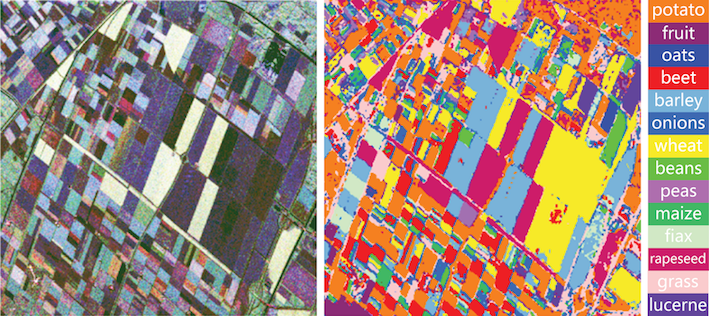}
\caption{Flevoland dataset. Left: Pauli RGB of the PolSAR dataset; Right: classification result from~\cite{ZZhang2017}.} \label{fig:polsar_classification}
\end{figure}

\par
\subsubsection{Terrain surface classification}
When terrain surface classification uses SAR, in particular polarimetric SAR (PolSAR), data is another important application in radar remote sensing. This is very similar to the task of image segmentation in computer vision. Conventional approaches are mostly based on pixel-wise polarimetric target decomposition parameters~\cite{JinandXu2013}. They hardly considered the spatial patterns, which convey rich information in high-resolution SAR images~\cite{XuGRSL2016}. Deep learning provides such a tool for automatically extract features that represent spatial patterns as well as polarimetric characteristics.
\par
One large stream of studies will employ at least one type of unsupervised generative graphical models, such as DBN, SAE or RBM.
\par
Xie \emph{et al.}~\cite{XieIGARSS2014} first introduce multi-layer feature learning for PolSAR classification, where SAE is employed to extract useful features from a  channel PolSAR image. 
\par
Geng \emph{et al.}~\cite{GengGRL2015} propose a deep convolutional autoencoder (DCAE) to extract features and conduct classification automatically. The DCAE consists of a hand-crafted first layer of convolution, which contains kernels, such as gray-level cooccurrence matrix and Gabor filters, and a hand-crafted second layer of scale transformation, which integrates correlated neighbor pixels. The rest layers are trained SAE. This approach is tested on high-resolution single-polarization TerraSAR-X images. Geng \emph{et al.}~\cite{GengTGRS2017} later propose a similar framework, called deep supervised and contractive neural network (DSCNN), for SAR image classification, which further includes the histogram of oriented gradient (HOG) descriptors as hand-crafted kernels. The trainable AE layers employ a supervised penalty, which captures the relevant information between features and labels, and a contractive restriction, which enhances local invariance. An interesting finding of Geng \emph{et al.}~\cite{GengTGRS2017} is that speckle reduction yields the worse performance and the authors suspect that speckle reduction might smooth out some useful information.
\par
Lv \emph{et al.}~\cite{Lv2015} test DBN on urban land use and land cover classification using PolSAR data. Hou \emph{et al.}~\cite{HouJSTARS16} propose SAE combined with superpixel for PolSAR image classification. Multiple layers of AE are trained on a pixel-by-pixel basis. Superpixels are formed based on Pauli-decomposed pseudo-color image. The output of SAE is used as a feature in the final step of k-nearest neighbor clustering of superpixels. Zhang \emph{et al.}~\cite{LZhangGRSL16} apply stacked sparse AE to PolSAR image classification. Qin \emph{et al.}~\cite{QinRSL17} apply adaptive boosting of RBMs to PolSAR image classification. Zhao \emph{et al.}~\cite{ZhaoPR17} propose discriminant DBN (DisDBN) for SAR image classification, in which the discriminant features are learned by combining ensemble learning with a deep belief network in an unsupervised manner.
\par
Jiao and Liu~\cite{JiaoTIP16} propose a deep stacking network for PolSAR image classification, which mainly takes advantage of fast Wishart distance calculation through linear projection. The proposed network aims to perform k-means clustering/classification task where Wishart distance is used as the similarity metric.
\par
The other stream of studies involves CNNs. Zhou \emph{et al.}~\cite{ZhouGRSL16} apply CNN to PolSAR image classification, where covariance matrix is extracted as 6-real-channel data input. Duan \emph{et al.}~\cite{DuanPR17} propose to replace the conventional-pooling layer in CNN by a wavelet-constrained pooling layer. The so-called convolutional-wavelet neural network is then used in conjunction with superpixels and Markov Random Field (MRF) to produce the final segmentation map.
\par
Zhang \emph{et al.}~\cite{ZZhang2017} propose a complex-valued (CV) CNN (cf. Fig.~\ref{fig:cv_cnn}) specifically designed to process complex values in PolSAR data, i.e., the off-diagonal elements of coherency or covariance matrix. CV-CNN not only takes complex numbers as input but also employs complex weights and complex operations throughout different layers. A complex-valued backpropagation algorithm is also developed to train it. Fig.~\ref{fig:polsar_classification} shows an example of PolSAR classification using CV-CNN.
\par
\par
\subsubsection{Parameter inversion}
Authors in~\cite{LWangTGRS2016} apply CNN to estimate ice concentration using SAR image during melt season. The labels are produced by visual interpretation by ice experts. It is tested on dual-pol RadarSat-2 data. Since the problem considered is regression of a continuous value, the loss function is selected as mean squared error. The final results suggest that CNN can produce a more detailed result then operational products.

\subsection{Interpretation of High-resolution Satellite Images}\label{sec:vhr}
\subsubsection{Scene Classification}

Scene classification, which aims to automatically assign a semantic label to each scene image, has been an active research topic in the field of high-resolution satellite images in the past decades~\cite{YangDTX12,ShaoYX2013,YangYX15,HUXWHZ2015,ZhaoZXZ16,HuXHZX16,Xia2015}. As a key problem in the interpretation of satellite images, it has widespread applications, including object detection~\cite{BhagavathyM06,ChengZH16}, change detection~\cite{Chen06}, urban planning, land resource management, etc. However, due to the high spatial resolutions, different scene images may contain the same kinds of objects or share similar spatial arrangement. For example, both residential area and commercial area may contain buildings, roads and trees, but they are two different scene types. Therefore, the great variations in the spatial arrangements and structural patterns make scene classification a considerably challenging task.
\par
Generally, scene classification can be divided into two steps: feature extraction and classification. With the growing number of images, to train a complicated nonlinear classifier is time-consuming. Hence, to extract a holistic and discriminative feature representation is the most significant part for scene classification. Traditional approaches are mostly based on the Bag-of-Visual-Words (BoVW) model~\cite{SivicZ03,ZhuZZXZ16}, but their potential for improvement was limited by the ability of experts to design the feature extractor and the expressive power encoded.

\par
The deep arhitectures discussed in Section~\ref{sec:cnns} have been applied to the problem of scene classification of high-resolution satellite images and led to
state-of-the-art performance~\cite{Zou2015,Xia2015,Penatti15,Castelluccio15,HUXWHZ2015,Hu20152,FZhang,Luus15,Nogueira15,Marmanis2016}. As deep learning is a multi-layer feature learning architecture, it can learn more abstract and discriminative semantic features as the depth grows and achieve far better classification performance compared with the mid-level approaches. In this section, we summarize the existing deep learning-based methods into the following three categories:
\begin{itemize}
\item[-] Using pre-trained networks. The pre-trained deep CNN on the natural image dataset, e.g., OverFeat~\cite{Ser14}, GoogLeNet~\cite{GoogLeNet}, etc., have led to impressive results on scene classification of high-resolution satellite images by directly extracting the features from the intermediate layers to form global feature representations~\cite{Penatti15,Castelluccio15,Hu20152,Marmanis2016}. For example, \cite{Xia2015,Penatti15} and \cite{Castelluccio15} directly use the features from the fully-connected layers as the input of the classifier, while~\cite{Hu20152} takes the CNN as local feature extractor and combine it with feature coding techniques, such as BoVW~\cite{SivicZ03} and Vector of Locally Aggregated Descriptors (VLAD) to generate the final image representation. 
\item[-] Making a pre-trained model adapt to the specific conditions observed in a dataset under study, one can decide to fine-tune it on a smaller labeled dataset of satellite images. For example, \cite{Castelluccio15} and \cite{Nogueira15} fine-tune some high-level layers of the GoogLeNet~\cite{GoogLeNet}, etc., using the UC-Merced dataset~\cite{UCM} (see Section~\ref{sec:RSD}), obtaining better results than directly using only the pre-trained CNNs. This can be explained because the features learned are more oriented to the satellite images after fine-tuning, which can help to exploit the intrinsic characteristic of satellite images. Nonetheless, compared with the natural image dataset that consists of more than ten millions of samples, the scales of public satellite image datasets (i.e., UC-Merced dataset~\cite{UCM}, RSSCN7 dataset~\cite{Zou2015}, WHU-RS19~\citep{Xiayang2010} are fairly small -- for example, up to several thousands -- for which we cannot fine-tune the whole CNNs to make them more adaptive to satellite images.
\item[-] Training new networks. In addition to the above two ways to use deep learning methods for classifying satellite images, some researchers train the network using satellite images from scratch. For example, \cite{Castelluccio15} and \cite{Nogueira15} train the networks by only using the existing satellite image dataset, which suffers a drop in classification accuracy compared with using the pre-trained networks as global feature extractors or fine-tuning the pre-trained networks. The reason may lies in the fact that the large-scale networks usually contain millions of parameters to be learned. Thus, training them using the small-scale satellite image datasets will easily cause overfitting and local minimum problems. Consequently, some construct a new smaller network and train it from scratch using satellite images to better fit the satellite images~\cite{Zou2015,FZhang,Luus15,Vol16}. However, such small-scale networks are often easily oriented to the training images, and the generalization ability decreases. For each satellite dataset, the network need to be retrained.
\end{itemize}

\subsubsection{Object Detection} Object detection is another important task in the interpretation of high-resolution satellite images~\cite{cheng2016survey}: one wishes to localize one or more specific ground objects of interest (such as building, vehicle, aircraft, etc.) within a satellite image and predict their corresponding categories 
as shown in Fig.~\ref{fig:object_detection}. 
Due to the powerful ability of learning high-level (more abstract and semantically meaningful) feature representations, the deep CNNs are being explored in object detection systems in opposition to the more traditional proposals methods followed by a classifier based on handcrafted features~\cite{yokoya2015object,han2014efficient}. Here, we review most existing works using CNNs for both specific and generic object detection.

\begin{figure}[tb!]
\centering
\includegraphics[width=0.45 \linewidth]{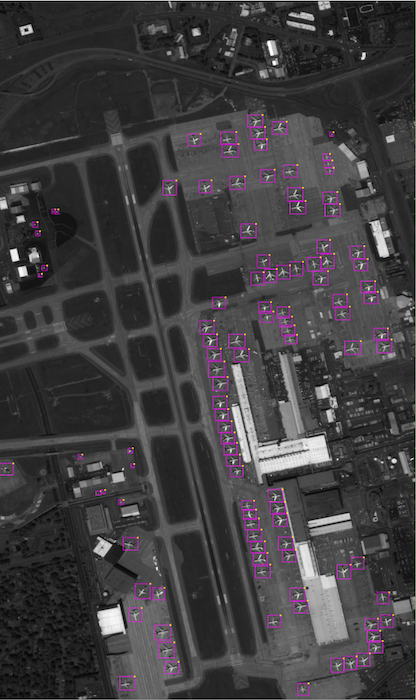}
\includegraphics[width=0.45 \linewidth]{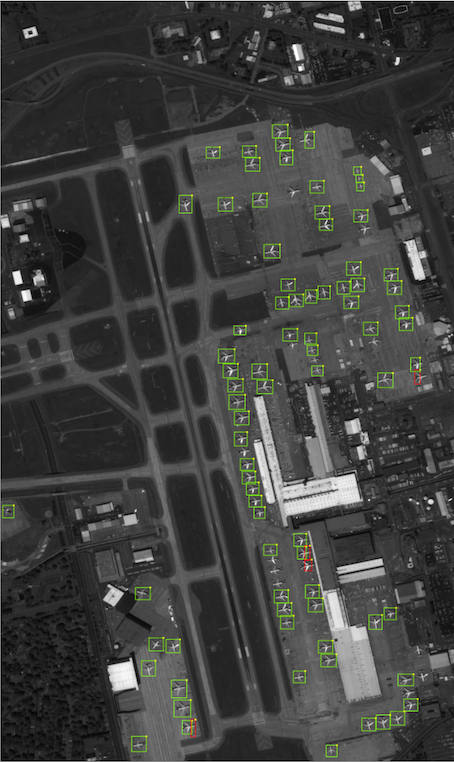}
\caption{Illustration of a typical object detection result within a high-resolution satellite image. The left: the annotated ground-truth of targets of interests (airplanes); The right: the airplanes detected by a CNN-based detector.} \label{fig:object_detection}
\end{figure}

Jin \emph{et al.}~\cite{jin2007vehicle} propose a vector-guided vehicle detection approach for IKONOS satellite imagery using a morphological shared-weight neural network, which learns the implicit vehicle model and incorporates both spatial and spectral characteristics, and classifies pixels into vehicles and non-vehicles.
To address the problem of large-scale variance of objects, Chen \emph{et al.}~\cite{chen2014vehicle} propose a hybrid deep CNN model for vehicle detection in satellite images, which divides all feature maps of the last convolutional and max-pooling layer of CNN into multiple blocks of variable receptive field size or pooling size, to extract multi-scale features. Jiang \emph{et al.}~\cite{jiang2015deep} propose a CNN-based vehicle detection approach, where a graph-based superpixel segmentation is used to extract image patches and a CNN model is trained to predict whether a patch contains a vehicle.

A few detection methods transfer the pre-trained CNNs for object detection. Zhou \emph{et al.}~\cite{zhou2016weakly} propose a weakly supervised learning framework to train an object detector, where a pre-trained CNN model is transferred to extract high-level features of objects and the negative bootstrapping scheme is incorporated into the detector training process to provide faster convergence of the detector. Zhang \emph{et al.}~\cite{zhang2015hierarchical} propose a hierarchical oil tank detector, which combines deep surrounding features, which are extracted from the pre-trained CNN model with local features (histogram of oriented gradients~\cite{dalal2005hog}). 
The candidate regions are selected by an ellipse and line segment detector. Salberg~\cite{salberg2015detection} proposes to extract features from the pre-trained AlexNet model and applies the deep CNN features for automatic detection of seals in aerial images. {\v{S}}evo \emph{et al.}~\cite{vsevo2016convolutional} propose a two-stage approach for CNN training and develop an automatic object detection method based on a pre-trained CNN, where the GoogLeNet is first fine-tuned twice on UC-Merced dataset, using different fine-tuning options, and then the fine-tuned model is utilized for sliding-window object detection. To address the problem of orientation variations of objects, Zhu \emph{et al.}~\cite{zhu2015orientation} employ the pre-trained CNN features that are extracted from combined layers and implement orientation-robust object detection in a coarse localization framework.

Zhang \emph{et al.}~\cite{zhang2016weakly} propose a weakly supervised learning approach using coupled CNNs for aircraft detection. The authors employ an iterative weakly supervised framework that simply requires image-level training data to automatically mine and augment the training data set from the original image, which can dramatically decrease human labor. A coupled CNN model, which is composed of a candidate region proposal network, and a localization network are developed to generate region proposals and locate the aircrafts simultaneously, which is suitable and effective for large-scale high-resolution satellite images.

For enhancing the performance of generic object detection, Cheng \emph{et al.}~\cite{ChengZH16} propose an effective approach to learn a rotation-invariant CNN (RICNN) to improve invariance to object rotation. In their paper, they add a new rotation-invariant layer to the off-the-shelf AlexNet model. The RICNN is learned by optimizing a new object function, including an additional regularization constraint which enforces the training samples before and after being rotating to share the similar features to guarantee the rotation-invariant ability of RICNN model.

Finally, several papers considering other methods than CNNs exist.
Tang \emph{et al.}~\cite{tang2015compressed} propose a compressed-domain ship detection framework combined with SDA and extreme learning machine (ELM)~\cite{huang2006extreme} for optical space-borne images. Two SDA models are employed for hierarchical ship feature extraction in the wavelet domain, which can yield more roust features under changing conditions. The ELM is introduced for efficient feature pooling and classification, making the ship detection accurate and fast. Han \emph{et al.}~\cite{han2015object} propose a effective object detection framework, exploiting weakly supervised learning and DBM. The system only requires weak label informing about the presence of an object in the whole image and significantly reduces the labor of manually annotating training data. 


\subsubsection{Image Retrieval} Remote sensing image retrieval aims at retrieving images with a similar visual content, with respect to a query image from a  database~\cite{yang2013geographic}. A common image retrieval system needs to compute image similarity based on image feature representations, and thus the performance of a retrieval depends on the descriptive capability of image features to a large degree. Building image  representation via feature coding methods (e.g., BoVW and VLAD) using low-level hand-crafted features has been proven to be very effective in aerial image retrieval~\cite{yang2013geographic,ozkan2014performance}. Nevertheless, the discriminative ability of low-level features is very limited, and thus it is difficult to achieve substantial performance gain. Recently, a few works have investigated extracting deep feature representations from CNNs. Napoletano~\cite{napoletano2016visual} extracts deep features from the fully-connected layers of the pre-trained CNN models, and the deep features prove to perform better than low-level features regardless of the retrieval system. Zhou \emph{et al.}~\cite{zhou2017learning} propose a CNN architecture followed by a three-layer perceptron, which is trained on a large remote sensing dataset and able to achieve remarkable performance even with low dimensional deep features. Jiang \emph{et al.}~\cite{JiangXia2017} present a sketch-based satellite image retrieval method by learning deep cross-domain features, which enables us to retrieve satellite images with hand-free sketches only.

Although there is still a lack of sufficient study of exploiting deep learning approaches for remote sensing image retrieval at present, in consideration of the great potential in learning high-level features of deep learning methods, we believe that more deep learning based image retrieval systems will be developed in the near future. It is also worth noticing that how to integrate the feedback from users into the deep learning retrieval scheme.

\subsection{Multimodal Data Fusion}\label{sec:df}
Data fusion is one of the fast-moving areas of remote sensing~\cite{Gom15,Sch16,dfc16}: due to the recent increases in availability of sensor data, the perspectives of using big and heterogeneous data to study environmental processes have become more tangible.
\par
Of course, when data are big and relations to be unveiled are complex, one would favour high capacity models: in this respect, deep neural networks are natural candidates to tackle the challenges of modern data fusion in remote sensing. Below, we review three areas of remote sensing image analysis where data fusion tasks have been approached with deep learning: pansharpening (Sec.~\ref{sec:PN}), feature and decision-level fusion (Sec.~\ref{sec:fuse}), and fusion of heterogeneous sources (Sec.~\ref{sec:hete}).

\subsubsection{Pansharpening and Super-Resolution}
\label{sec:PN}
Pansharpening is the task of improving the spatial resolution of multispectral data by fusing it with data characterized by sharper spatial information. It is a special instance of the more general problem of super-resolution. Traditionally, the field was dominated by works fusing multispectral data with panchromatic bands~\cite{Alp07}, but more recently it has been extended to thermal~\cite{fas07} or hyperspectral images~\cite{Lonc15}. Most techniques rely either on projective methods, sparse models, or pyramidal decompositions. Using deep neural networks for pansharpening multispectral images is certainly an interesting concept, since most image acquired by satellite as the WorldView series or Landsat come with a panchromatic band. In this respect, training data are abundant, which is in line with the requirements of modern CNNs.
\par
A first attempt in this direction can be found in~\cite{Zho16}, where authors use a shallow network to upsample the intensity component obtained after the IHS of color images (RGB). Once the multispectral bands have been upsampled with the CNN, a traditional Gram-Schmidt transform is used to perform the pansharpening. The authors use a dataset of QuickBird images for their analysis. Even though this is interesting, in this paper, authors simply replace one operation (the nearest neighbours or bicubic convolution) with a CNN.
\par
In~\cite{Mas16}, authors propose using a CNN to learn the pansharpening transform end-to-end, i.e., letting the CNN perform the whole pansharpening process. In their CNN, they stack upsampled spectral bands with the panchromatic band and then learn, for each patch, the high resolution values of the central pixel.

In~\cite{Yua17} authors use a super-resolution CNN trained on natural images~\cite{Don14} as a pre-trained model and fine-tune it on a dataset of hyperspectral images. By doing so, they make an attempt at transfer leaning~\cite{Tui15b} between the domains of color (three bands, large bandwidths) and hyperspectral images (many bands, narrow bandwidths). Fine-tuning existing architectures, which have been trained on massive datasets with very big models, is often a relevant solution, since one makes use of discriminative strong features and only injects task-specific knowledge.
\par
In \cite{Hua15}, authors learn an upsampling of the panchromatic band via a stack of autoencoders: the model is trained to predict the full-resolution panchromatic image, from  a downsampled version of itself (at the resolution of the multispectral bands). Once the model is trained, the multispectral bands are fed into the model one by one, therefore being upsampled using the data relationships learned from the panchromatic images.

\subsubsection{Feature and Decision-level Fusion for {image classification}}
\label{sec:fuse}
Most current remote sensing literature, dealing with deep neural networks, studies the problem of {\emph{image  classification}}, i.e. the task of assigning each pixel in the image to a given semantic class (land use, land cover, damage level, etc.). 
In the following, we review recent approaches dealing with {image classification} problems, mostly at very-high resolution, using two strategies: \emph{feature-level} fusion and \emph{decision-level} fusion. In the last part of this section, we will also review works using different data sources to tackle separate, but related predictive tasks, or \emph{multi-task} problems.
\par
i) Feature-level fusion: using multiple sources simultaneously in a network. As most image processing techniques, deep neural networks use $d$-dimensional inputs. A very simple way of using multiple data sources in a deep network is to \emph{stack} them, i.e., to concatenate the image sources into a single data cube to be processed. The filters learned by the first layer of the network will, therefore, depend on a stack of different sources. Studies considering this straightforward extension of neural networks are numerous and, in~\cite{Lag15}, authors compared networks trained on color RGB data (fine tuned from existing architectures) with networks, including a DSM channel on the 2015 Data Fusion Contest dataset over the city of Zeebruges~\cite{DFCA}\footnote{Data are available at \url{http://www.grss-ieee.org/community/technical-committees/data-fusion/}{, also see Section~\ref{sec:RSD}}}. They use the CNN as a feature extractor and then use the features to train a SVM, predicting a {single} semantic class {for the entire} patch. They then apply the classifier in a sliding window manner.
\par
Parallel research considered spatial structures in the network, by training architectures predicting all labels in the patch, instead of a single label to be attributed to the central pixel. By doing so, spatial structures are inherently included in the filters. Fully convolutional and deconvolutional approaches are natural candidates for such task: in the first, the last fully connected layer is replaced with a convolutional layer (see~\cite{Ser14}) to have a downsized patch prediction that then needs to be upsampled. In the second, a series of deconvolutions (transposed convolutions~\cite{Deconv4Seg2015,FCN}) are learned to upsample the convolutional fully connected layer. Both approaches have been compared in~\cite{Vol16} on the ISPRS Vaihingen and Potsdam benchmark datasets\footnote{Available at \url{http://www2.isprs.org/commissions/comm3/wg4/semantic-labeling.html}{, also see Section~\ref{sec:RSD}}} stacking color infrared (CIR: infrared, red and green channels) and a normalized digital elevation model. The architectures compared and some zoomed results are reported in Figs.~\ref{fig:mitcharchi} and~\ref{fig:mitchres}, respectively. Other strategies to spatial upsampling have been proposed in the recent literature, including the direct use of upsampled activation maps as features to train the final classifier~\cite{Mag17}. In~\cite{Kam16}, authors studied the possibility of visualizing uncertainty of predictions (applying the model of~\cite{Gal15}): they stacked CIR, DSM and normalized DSM data as inputs to the CNN.

\begin{figure*}[tbh!]
\centering
\includegraphics[width=0.8\linewidth]{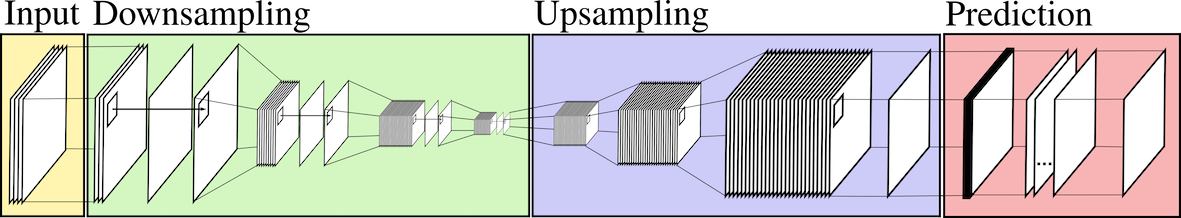}
\caption{Deconvolution network proposed in~\cite{Vol16}. The yellow and green part correspond to a fully convolutional network with a $9 \times 9$ pixels bottleneck; then a deconvolutional block (purple) leads to predictions of the same size as the input image (in~\cite{Vol16}, $65 \times 65$ pixels).\label{fig:mitcharchi}}
\end{figure*}

\begin{figure*}[!ht]
\centering
\begin{tabular}{rccc|ccc}
 & {\bf Image} & {\bf nDSM} & {\bf GT} & {\bf CNN-PC} & {\bf CNN-SPL} & {\bf CNN-FPL} \\

&
\includegraphics[width=.12\textwidth]{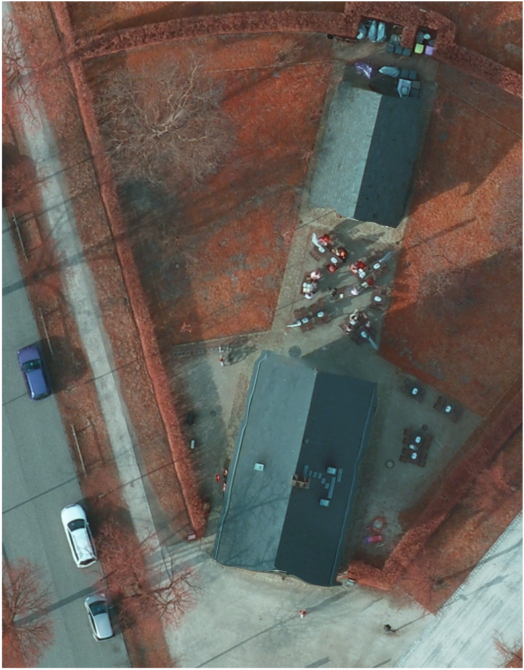} &
\includegraphics[width=.12\textwidth]{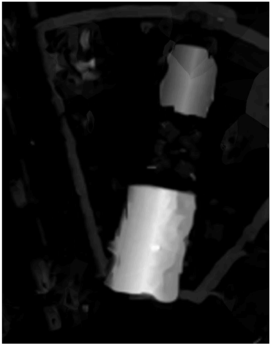} &
\includegraphics[width=.12\textwidth]{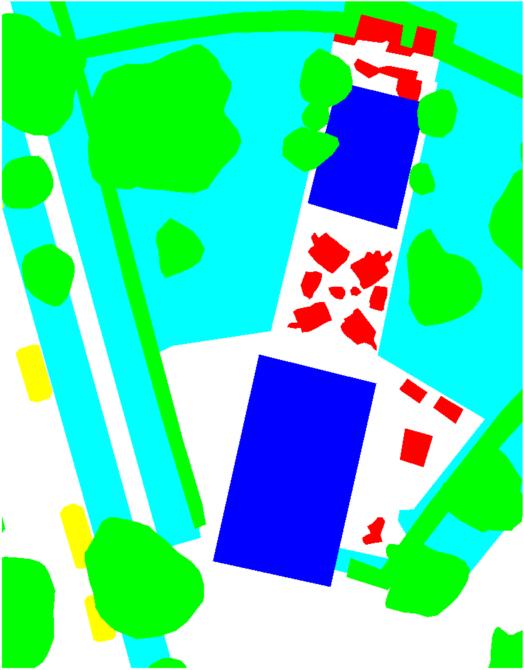} &
\includegraphics[width=.12\textwidth]{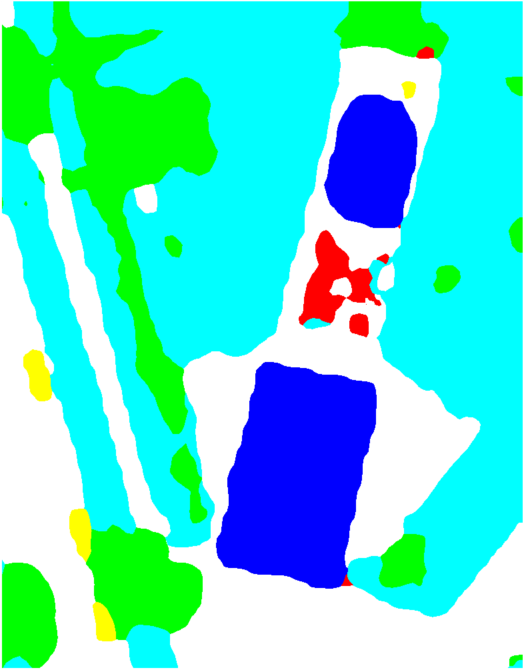} &
\includegraphics[width=.12\textwidth]{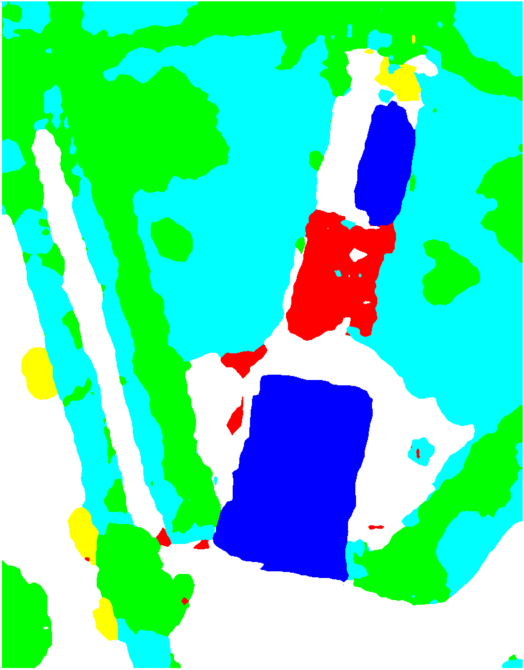} &
\includegraphics[width=.12\textwidth]{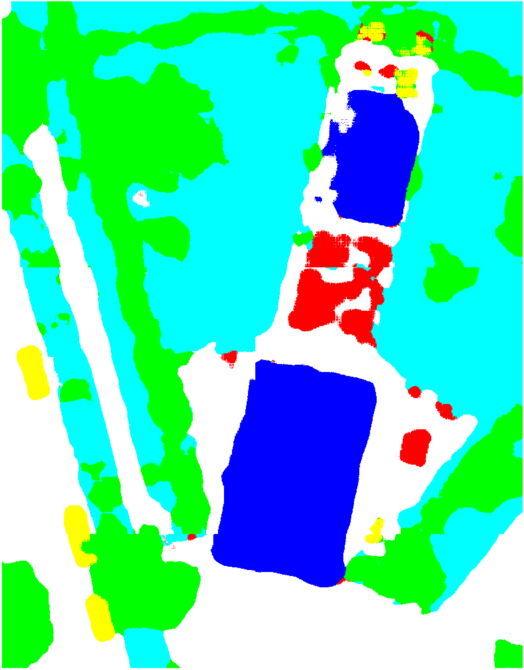} \\[-0.75mm]

 &
\includegraphics[width=.12\textwidth]{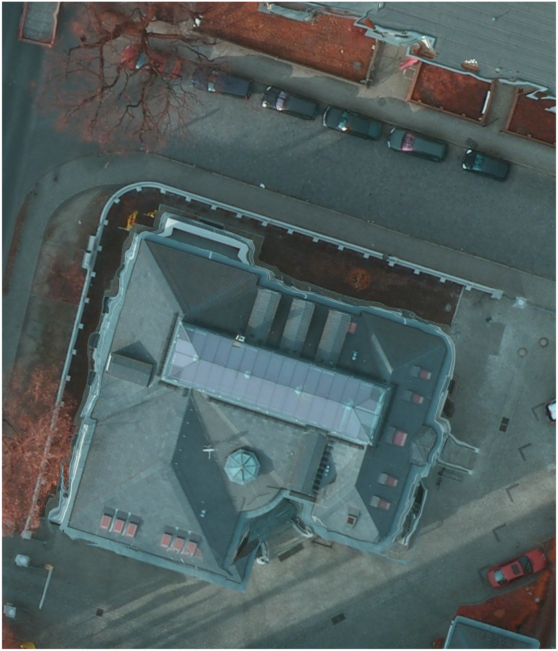} &
\includegraphics[width=.12\textwidth]{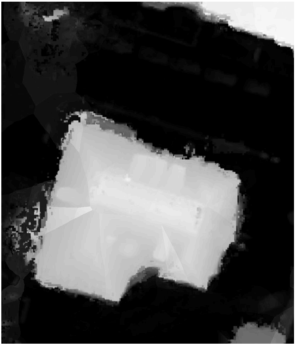} &
\includegraphics[width=.12\textwidth]{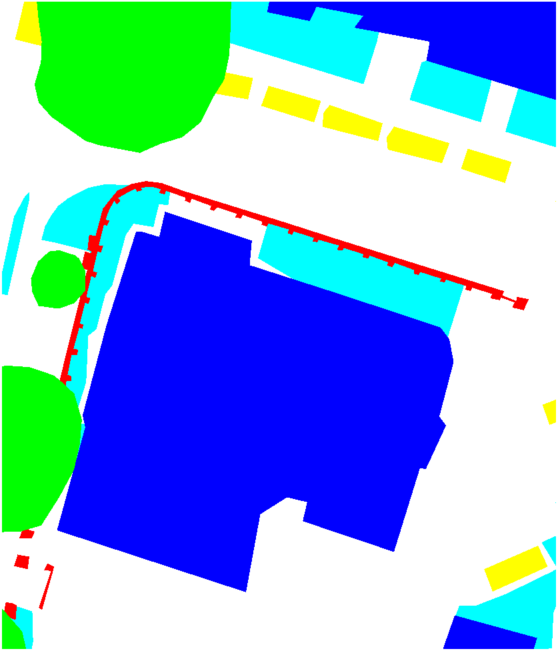} &
 \includegraphics[width=.12\textwidth]{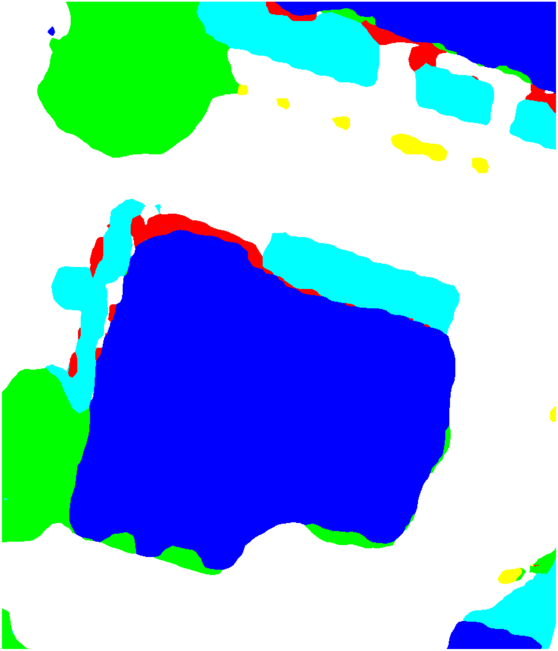} &
 \includegraphics[width=.12\textwidth]{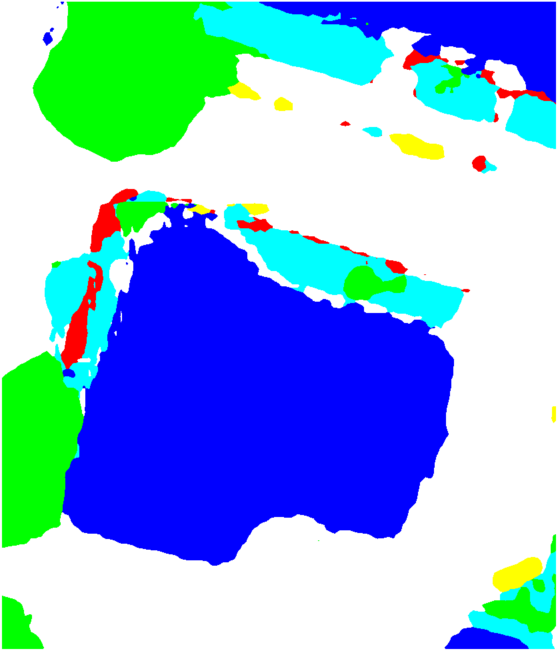} &
  \includegraphics[width=.12\textwidth]{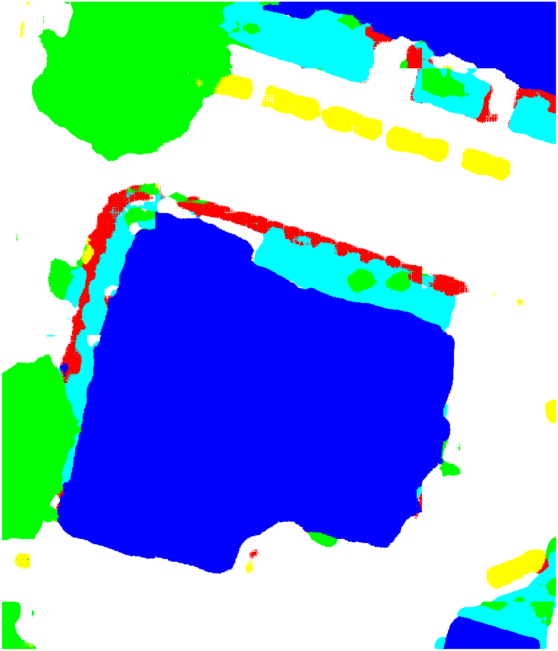} \\[-0.75mm]

 &
\includegraphics[width=.12\textwidth]{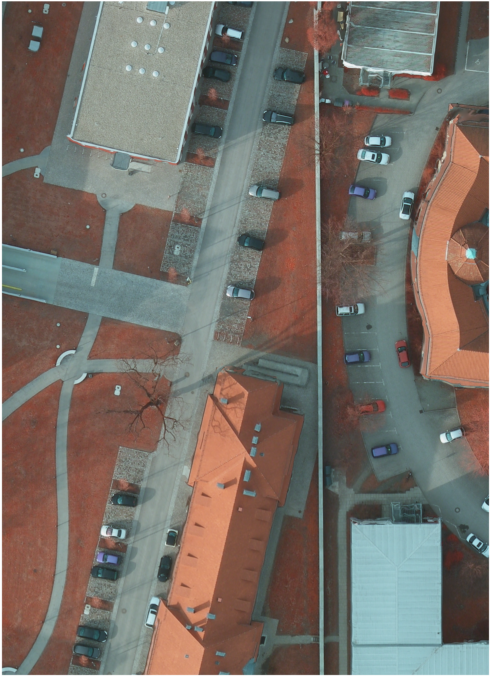} &
\includegraphics[width=.12\textwidth]{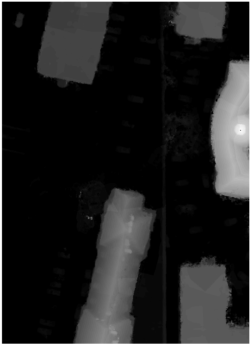} &
\includegraphics[width=.12\textwidth]{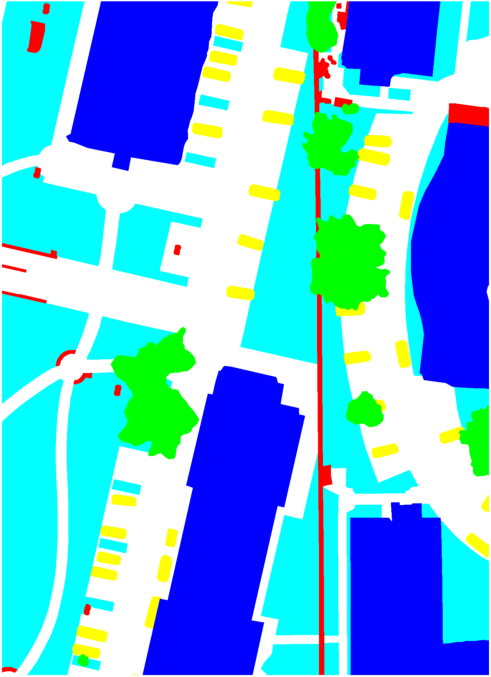} &
 \includegraphics[width=.12\textwidth]{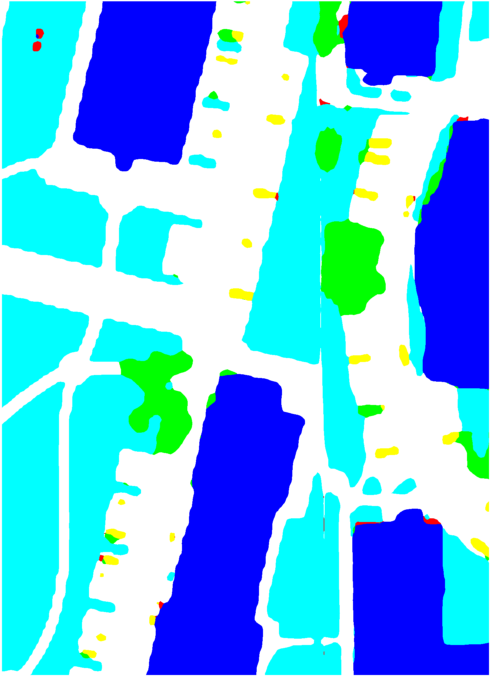} &
  \includegraphics[width=.12\textwidth]{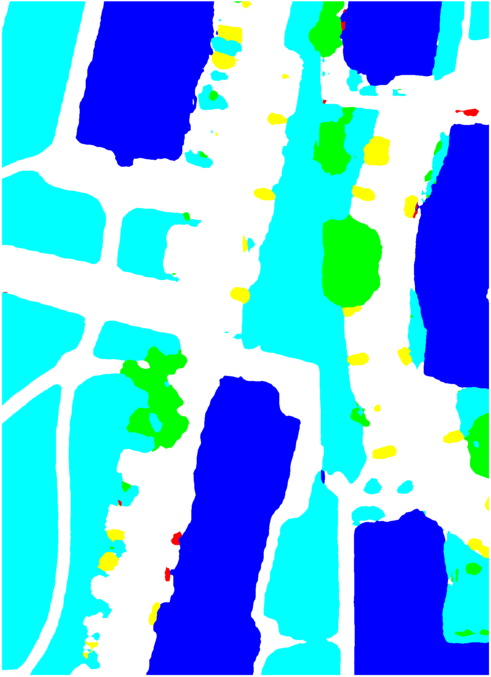} &
   \includegraphics[width=.12\textwidth]{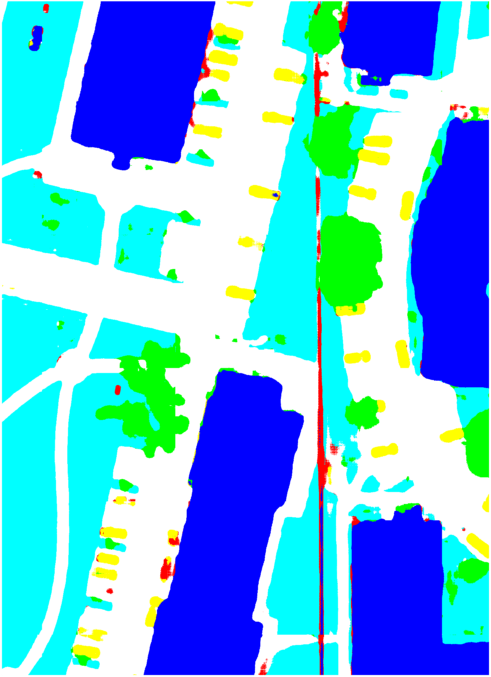} \\[-0.75mm]

\end{tabular}
\caption{{Image classification} results on the Potsdam datasets, considering $65 \times 65$ pixels patches (from~\cite{Vol16}). \textbf{CNN-PC}: patch-based CNN, predicting single labels per patch and using a sliding window approach; \textbf{CNN-SPL}: fully convolutional CNN, predicting a $9\times9$ output, then upsampled to the original size via interpolation; \textbf{CNN-FPL}: deconvolutional network predicting the $65 \times 65$ output at full resolution. \label{fig:mitchres}}
\end{figure*}

Besides dense predictions, other strategies have been proposed to include spatial information in deep neural networks: for example, authors of~\cite{GengTGRS2017} extract different types of spatial filters and stack them in a single tensor; the tensor is then used to learn a supervised stack of autoencoders. They apply their models on the {classification} of SAR images, so the fusion here is to be considered between different types of spatial information. The neural network is then followed by a conditional random field, to decrease the effect of speckle noise inherent in SAR images. In~\cite{Tui15}, authors learn combinations of spatial filters extracted from hyperspectral image bands and DSMs: even if the model is not a traditional deep network, it learns a sequence of recombinations of filters, extracting therefore higher level information in an automatic way as deep neural networks do. {I.e.}, it learns the right filters parameters (along with their combinations) instead of learning the filters coefficients themselves.
\par
Data fusion is also a key component in change detection, where one would like to extract joint features from a bi-temporal sequence. The aim is to learn a joint representation, where both (co-registered) images can be compared: this area is especially interesting when methods can align data from multiple sensors (see~\cite{Vol14b,Mar16}). Three studies employ deep learning to this end:
\begin{itemize}
\item[-] \cite{Gon17}, where authors learn a joint representation of two images with Deep Belief Networks. Feature vectors issued from the two image acquisitions are stacked and used to learn a representation, where changes stand out more clearly. Using such representation, changes are more easily detected by image differencing. Their approach is applied on optical images from the Chinese GaoFen-1 satellite and WorldView-2.
\item[-] \cite{Zha16}, where the joint representation is learned via a stack of autoencoders using the single temporal acquisitions at each end of the encoder-decoder system. By doing so, they learn a representation useful for change detection at the bottleneck of the system (i.e. in the middle). The authors show the versatility of their approach by applying it to several datasets, including pairs of optical and SAR images and an example performing change detection between optical and SAR images.
\item[-] More recent work, instead, addresses the transferability of deep learning for change detection, while analyzing data of long time series for large-scale problems. For example, in~\cite{Lyu}, authors make use of an end-to-end RNN to solve the multi/hyper-spectral change detection task, since RNN is well known to be good at processing sequential data. In their framework, an RNN based on long short-term memory (LSTM) is employed to learn joint spectral feature representations from a bi-temporal image sequence. In addition, authors also show that their network can detect multi-class changes and has a good transferability for change detection in a new scene without fine-tuning. Authors of~\cite{Lyu17} introduce a RNN-based transfer learning approach to detect annual urban dynamics of four cities (Beijing, New York, Melbourne, and Munich) from 1984 to 2016, using Landsat data. The main challenge here is that training data in such large-scale and long-term image sequence are very scarce. By combining RNN and transfer learning, they are able to transfer the feature representations learned from few training samples to new target scenes directly. Some zoomed results are reported in Fig.~\ref{fig:munich_change}.

\end{itemize}

\begin{figure*}[htb!]
\centering
\includegraphics[width=0.8\textwidth]{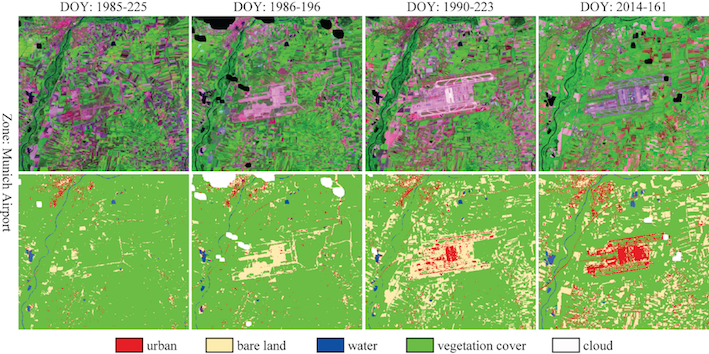}
\caption{Using large-scale and long-term multi-temporal image sequence, a deep learning-based system helps us to analyze how land cover changes. This example shows how Munich airport was built over the past 30 years.
\label{fig:munich_change}}
\end{figure*}

Another view on feature fusion can be found when considering neural networks fusing features obtained from different inputs: two (or more) networks are trained in parallel and their activations are then fused at a later stage, for example by feature concatenation. The author of~\cite{She16} studies a solution in this direction which fuses two CNNs: the first considers CIR images of the Vaihingen dataset and passes them through the pre-trained VGG network to learn color features, while the second considers the DSM and learns a fully connected network from scratch. Both models' features are then concatenated and two randomly initialized fully connected layers are learned from this concatenation. A similar logic is also followed in~\cite{Mar16b}, where authors learn a fully connected layer performing the fusion between networks learned at different spatial scales. They apply their model on tasks of buildings and road detection. In~\cite{HuJURSE17}, authors train a two-stream CNN with two separate, yet identical convolutional streams, which process the PolSAR and hyperspectral data in parallel, and only fuse the resulting information at a later convolutional layer for the purpose of land cover classification. With similar network architecture and contrastive loss function, authors of~\cite{MouJURSE17} learn a network for the identification of corresponding patches in SAR and Optical imagery of urban scenes.

ii) Decision-level fusion: fusing CNN (and other) outputs. If the works reviewed above use a single network to learn the semantics of interest all at once (either by extracting relevant features or by learning the model end to end), another line of works has studied ways of performing decision fusion with deep learning. Even though the distinction with the models reviewed above might seem artificial, we review here approaches including an explicit fusion layer between {land cover} maps. We distinguish between two families of approaches, depending on whether the decision fusion is performed as a post-processing step or learned:
\begin{itemize}
\item[-] \emph{Fusing semantic maps obtained with CNNs}: in this case, different models predict the {classes} and their predictions are then fused. Two works are particularly notable in this respect: on the one hand~\cite{Pai16}, authors fuse a {classification} map obtained by a CNN with another obtained by a random forest classifier which is trained using hand-crafted features. Both models use CIR, DSM and normalized DSM inputs from the Vaihingen dataset. The two  maps are fused by multiplication of the posterior probabilities and an edge-sensitive CRF is also learned on top to improve the quality of the final labeling. On the other hand, authors in~\cite{Mar16} consider learning an ensemble of CNNs and then averaging their predictions: their proposed pipeline has two main streams, one processing the CIR data and another processing the DSM. They train several CNNs, using them as inputs the activation maps of each layer of the main model, as well as one fusing the CIR and DEM main streams as in~\cite{She16}. By doing so, they obtain a series of {land cover} maps to nourish the ensemble, which improves performances by considering classifiers issued form different data sources and levels of abstraction. Compared to the previous one in this section, this model has the advantage of being entirely learned in an end-to-end fashion, but also incurs in extreme computational load and a complex architecture involving many skip connections and fusion layers.
\item[-] \emph{Decision fusion learned in the network}: an alternative to an ad-hoc fusion (multiplication or averaging of the posterior maps), one could learn the optimal fusion. In~\cite{Aud16}, authors perform the fusion between two maps obtained by pre-trained models by learning a fusion network  based on residual learning~\cite{ResNet} logic: in their architecture, they learn how to correct the average fusion result by learning extra coefficients favouring one or the other map. Their results show that such a learned fusion outperforms the, yet more intuitive, simple averaging of the posterior probabilities.

\end{itemize}

iii) Using CNN for solving different tasks. So far, {only literature dealing with a single task (image classification) has been reviewed}. But, besides it, one might want to predict other quantities or use the {image classification} results to improve the quality of related tasks as image registration. In this case, predicting different outputs jointly allows one to tighten feature representations with different meaning{s}, therefore leading to another type of data fusion with respect to the one seen until now (that was mainly concerned by fusing different inputs). Summarizing, we will talk about \emph{fusing outputs}. Below, we discuss three examples from recent literature, where alternative tasks are learned together with {image classification}.
\begin{itemize}
\item[-] \emph{Edges.} In the last section, we discussed the work of Marmanis \emph{et al.}~\cite{Mar16}, where authors were producing and fusing an ensemble of {land cover} maps. In~\cite{Mar17}, that work was extended by including the idea of predicting \emph{object boundaries} jointly with the {land cover}.  The intuition behind this is that predicting boundaries helps to achieve sharper (and therefore more desirable) classification maps. In~\cite{Mar17}, authors learn a CNN to separately output edges likelihoods at multiple scales from CIR and height data. Then, the boundaries detected with each source are added as an extra channel to each source and an {image classification} network, similar to the one in~\cite{Mar16} is trained. The predictions of such model are very accurate but the computational load involved becomes very high: authors report models involving up to $800$ millions of parameters to be learned.

\item[-] \emph{Depth.} Most approaches discussed above include the DSM as an input to the network. But, often, such information is not available (and it is certainly not when working on historical data). A system predicting a height map form image data would indeed be very valuable, since it could generate reasonably accurate DSM for color image acquisitions. This is known in vision as the problem of estimating depth maps~\cite{Eig14} and has been considered in~\cite{Sri17} for monocular subdecimeter images. In their models, authors use a joint loss function, which is a linear combination of a {dense image classification} loss and a regression loss minimizing DSM predictions errors. The model can be trained by traditional back-propagation by alternating over the two losses. Note that, in this case, the DSM is used as an output (contrarily with most approaches above) and is, therefore, not needed at prediction time.

\item[-] \emph{Registration.} When performing change detection, one expects perfect co-registration of the sources. But, especially when working at very high resolution, this is difficulty to achieve. For instance, think of urban areas, where buildings are tilted by the viewing angle. In their entry to the IEEE GRSS data fusion contest 2016\footnote{Data are available on \url{http://www.grss-ieee.org/community/technical-committees/data-fusion/}}, authors of~\cite{Vak16b} learn jointly the registration between the images, the {land use classification} of each input and a change detection map with a conditional random field model. The {land use} classifier used is a two-layers CNN trained from scratch and the model is applied successfully either to pairs of VHR images or to datasets composed of VHR images and video frames from the International Space Station.

\end{itemize}

\subsubsection{Fusing Heterogeneous Sources}
\label{sec:hete}
Data fusion is not only about fusing image data with the same viewpoint. Multimodal  remote sensing data exceed these restrictive boundaries and approaches to tackle new, exciting problems with remote sensing are appearing in the literature. A strong example is the joint use of ground-based and aerial images~\cite{Lef17}: services, such as Google Street View and Flickr, provide endless sources of ground images describing cities from the human perspective. These data can be fused to overhead views to provide better object detection, localization, or recreation of virtual environments. In the following, we review a series of applications in this direction.
\par
In~\cite{Wegner2016}, authors consider the task of detecting and classifying urban trees. To this end, they exploit Faster R-CNN~\cite{Ren15}, an object detector developed for general purpose object detection in vision. After detecting the trees in the aerial view and the Google Street View panoramas, they minimize a energy function to detect trees jointly in all sources, but also avoiding multiple and illogical detections (e.g. trees in the middle of a street lane). They use a trees inventory from the city of Pasadena to validate their detection model and train a fine grained CNN based on GoogleNet~\cite{GoogLeNet}, to perform fine-grained classification of the trees species on the detections, with impressive results. Authors of~\cite{Mattyus16} take advantage of an approach that combines CNN and MRF and can estimate fine grained categories (e.g., road, sidewalk, background, building and parking) by doing joint inference over both monocular aerial imagery and ground images taken from a stereo camera on top of a car.
\par
Many papers in geospatial computer vision work towards cross-view image localization: when presented to a ground picture, it would be relevant to be able to locate it in space.  This is very important for photo-sharing platforms, for which only a fraction of the uploaded photos comes with geo-location. Authors of~\cite{lin2015learning,Vo16} worked towards this aim, by training a cross-view Siamese network~\cite{Cho05} to match ground images and aerial views. Siamese networks have also been recently applied to detect changes between matched ground panoramas and aerial images in~\cite{Lef17}. Back to more traditional CNNs, authors of~\cite{Wor15,Wor15b} study the specificity of images to refer to a given city: they study how much images of Charleston resemble those from San Francisco, and the other way around, by using the fully connected layers of Places CNN~\cite{Zho14} and then translating this into differences in the respective aerial images. Moreover, in~\cite{Wor15} they also present applications on image localization similar to the above, where the likelihood of localization is given by a similarity score between the features of the fully connected layer of Places CNN.
\par

\subsection{3D Reconstruction}\label{sec:3D}

3D data generation from image data plays an important role for remote sensing. 3D data (e.g., in form of a DSM or DTM) is a basic data layer for further processing or analysis steps. The processing of image data from airborne sensors or satellite systems is a long-standing tradition. In a typical 3D data generation workflow, two main steps must be performed. First, camera orientation, which means computing the position and orientation of the cameras that produced the image. This can be computed from the image data itself, by identifying and matching tie-points and then performing camera-resectioning. The second step is triangulation with calculates the 3D measurements for point correspondences that get established through stereo matching.
The fundamental algorithms in this pipeline are of geometrical nature, the implementations are based on analytical calculations. So far, machine learning did not play a big role in this pipeline. However, there are steps in this pipeline which recently could be improved significantly by using machine learning techniques.
\par
\subsubsection{Tie points identification and matching}
For instance, during camera orientation, the identification and matching of tie-points has long been done manually by operators. The task of the operator was to identify corresponding locations in two or more images. This process has been automated by clever engineering of computer algorithms to detect point locations in images that will be easy to re-detect in other images (e.g. corners) as well as algorithms for computing similarities of image patches for finding a tie-point correspondence. Many different detectors and similarity measures have been engineered so far, famous examples are the SIFT~\cite{Lowe2004} or SURF~\cite{Bay2008} features. However, all these engineered methods fall short of the last mile (i.e., they are still less accurate than humans). This is a domain where machine learning and, in particular, convolutional neural networks are employed to learn, based on a huge amount of correct tie-point matches and point locations what characterizes tie-points and what is the best way of computing the similarity between them is.
In the area of tie-point detection and matching, Fischer \emph{et al.}~\cite{FischerDB14} used CNN to learn a descriptor for image patch matching from training examples, similar to the well-known SIFT descriptor. In this work, authors trained a CNN with 5 convolutional layers and 2 fully connected layers. The trained network computes a descriptor for a given image patch. In the experiments on standard data sets, authors could show that the trained descriptors outperform engineered descriptors (i.e. SIFT) significantly in an tie-point matching task. Similar successes are described in other works by Handa \emph{et al.}~\cite{handa2016}, Lenc and Vedaldi~\cite{lenc2016}, and Han \emph{et al.}~\cite{Han2015} The work of Yi \emph{et al.}~\cite{Yi16} takes this idea one step further: the authors proposed a deep CNN to detect tie-point locations in an image and output a descriptor vector for each tie-point.

\subsubsection{Stereo processing using convolutional neural networks}
The second important step in this workflow is stereo matching, i.e., the search for corresponding pixels in two or more images. In this step, a corresponding pixel is sought for every pixel in the image. In most cases, this search can be restricted to a line in the corresponding image. However, current methods still make mistakes in this process.  The semi-global matching (SGM) approach by Hirschmueller~\cite{hirschmueller2016} acted as the gold-standard method for some considerable time.

Since 2002, the progress on stereo processing is tracked by the Middlebury stereo evaluation benchmark\footnote{\url{http://vision.middlebury.edu/stereo/}}. The benchmark allows to compare results of stereo processing algorithms to a carefully maintained ground truth. The performance of the different algorithms can be viewed as a ranked list. This ranking reveals that, today, the top performing method is based on CNNs.

Most stereo methods in this ranking proceed along the following main steps. First, a stereo correspondence search is performed by computing a similarity measure between image locations. This is typically done exhaustively for all possible depth values. Next, the optimal depth values are searched by optimization on the cost value. Different optimization schemes, convex optimization, local-optimization strategies (e.g. SGM), and probabilities methods (e.g. MRF inference) are used. Finally, typically some heuristic filtering is applied to remove gross outliers (e.g. left-right check).

The pioneering work of Zbontar and LeCun~\cite{zbontar} utilized a CNN in the first step of the typical stereo pipeline. In their work, the authors proposed to train a CNN to compute the similarity measure between image patches (instead of using NCC or the Census transform). This change proved to be significant. Compared to SGM, which is often considered as a baseline method the proposed method achieved a significantly lower error rate. For SGM the error rate was still 18.4\% while for the MC-CNN method the error rate was only 8\%. After that, other variants of CNN-based stereo methods have been proposed and the best ranking method, today, has an error rate of only 5.9\%. In table~\label{tab:middlebury} the error rates of the top-ranking CNN-based methods are listed. 

\begin{table}
\centering
\caption{Top ranking stereo methods from the Middlebury stereo evaluation benchmark as of May 2017. CNN based methods are leading the board.}  \label{tab:middlebury}
\begin{tabular}{cc}
\hline
Method & bad pixel error rate \% \\
3DMST~\cite{Li:17} & 5.92\\
MC-CNN+TDSR~\cite{Drouyer2017} & 6.35 \\
LW-CNN~\cite{park2016} & 7.04 \\
MC-CNN-acrt~\cite{zbontar} & 8.08\\
SGM~\cite{hirschmueller2016} & 18.4\\
\hline
\end{tabular}
\end{table}

In addition to similarity measures, a typical stereo processing pipeline contains other engineered decisions as well. After creating a so-called cost volume from the similarity measures, most methods use specifically engineered algorithms for finding the depths (e.g. based on neighborhood constraints) and heuristics to filter out wrong matches.
New proposals however, suggest that these other steps can also be replaced solely by a CNN. Mayer \emph{et al.}~\cite{MIFDB16} proposed such a paradigm shifting design for stereo processing. In their proposal, the stereo processing problem is solely modeled as a CNN. The proposed CNN takes two images of a stereo pair as and input and directly outputs the final disparity map. A single CNN architecture replaces all the individual algorithms steps utilized so far. The CNN of Mayer is based on an encoder-decoder architecture with a total of 26 layers. In addition it includes crosslinks between contracting and expanding network parts. To train the CNN architecture, end-to-end training using ground truth image-depth map pairs is performed. The fascinating fact of the proposed method is that the stereo algorithm itself can be learned from data only. The network architecture itself does not define the algorithm but the data and the end-to-end training defines what type of processing the network should perform.

\subsubsection{Large scale semantic 3D city reconstruction}
The availability of semantics (e.g. the knowledge of what type of object a pixel in the image represents) through CNN-based classification is also changing the way that 3D information is generated from image data. The traditional 3D generation process did neglect object information. 3D data was generated from geometric constraints only. Image data were treated as pure intensity values without any semantic meaning. The availability of semantic information from CNN-based classification now makes it possible to utilize this information in the 3D generation process. CNN-based classification allows one to assign class labels to aerial imagery with unprecedented accuracy\cite{Mrmns2016}. Pixels in images are then assigned labels like vegetation, road, building etc. This semantic information can now be used to steer the 3D data generation process. Class label specific parameters can be chosen for the 3D data generation process. 

The latest proposal in this area, however, is a joint reconstruction of 3D and semantic information. This has been proposed in the work of Haene \emph{et al.}~\cite{haene2013}, where 3D reconstruction is performed with a volumetric method. The area to be reconstructed is partitioned into small cells, the size of it defining the resolution of the 3D reconstruction. The reconstruction algorithm now finds the optimal partitioning of this voxel grid into occupied and non-occupied voxels which fits to the image data. The result is a 3D reconstruction of the scene. The work of Haene \emph{et al.} also jointly assigns the 3D reconstruction to a class label for each voxel, e.g. vegetation, building, road, sky. Each generated 3D data point now also  has a class label. The 3D reconstruction is semantically interpretable. This process is a joint process, the computation of the occupied and non-occupied voxels takes into account the class labels in the original images. If a voxel corresponds to a building pixel in the image, it is set to occupied with high probability. If a voxel corresponds to a sky pixel in the image, it has a high probability of being unoccupied. On the other hand, if a set of voxels are stacked on top of each other, it is likely that these belong to some building, i.e. the probability for assigning the label class of building is increased for this structure.
This semantic 3D reconstruction method has been successfully applied to 3D reconstruction from aerial imagery by Blaha \emph{et al.}~\cite{blaha2016,Blh2016}. In their work they achieved a semantic 3D reconstruction of cities on large scales. The 3D model not only contains 3D data but also class labels. E.g., 3D structure that represents buildings gets the class label of building. Even more, every building has even the roof structures labeled as roof. Fig.~\ref{fig:Semantic3DBlaha} shows an image of a semantic 3D reconstruction produced by the method of ~\cite{blaha2016}.

\begin{figure}[htb!]
\centering
\includegraphics[width=0.85\columnwidth]{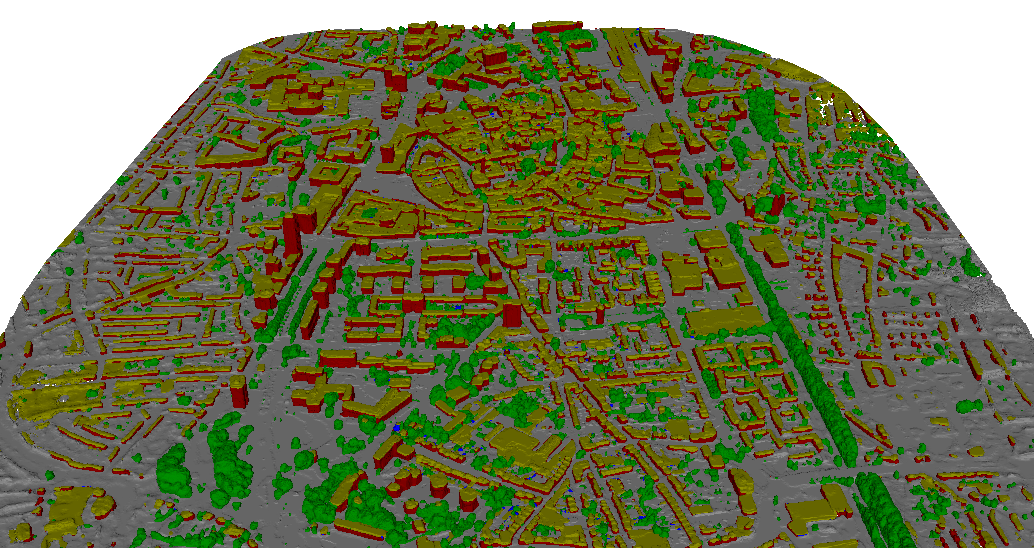}
\caption{Semantic 3D reconstruction from the Enschede aerial image data set computed with the method of~\cite{blaha2016}. The different colors represent different class labels: ground (gray), building (red), roof (yellow), vegetation (green) and clutter (blue). (Figure provided by the authors of~\cite{blaha2016}) \label{fig:Semantic3DBlaha}}
\end{figure}

In summary, it can be said that CNNs quickly took on a significant role in 3D data generation. Utilizing CNNs for stereo processing significantly boosted the accuracy and precision of depth estimation. On the other hand, the availability of reliable class labels extracted from CNNs classifiers opened the possibility of creating semantic 3D reconstructions, a research areas which is about to grow significantly.

\section{Deep Learning in Remote Sensing Made Ridiculously Simple to Start with}\label{sec:tools}
To make an easy start for researchers who attempt to work on deep learning in remote sensing, we list some available resources, including tutorials (Sec.~\ref{sec:Tut}) and open-source deep learning frameworks (Sec.~\ref{sec:ODLF}). In addition, we provide a selected list of open remote sensing data for training deep learning models (Sec.~\ref{sec:RSD}), as well as some showcasing examples with source codes developed using different deep learning frameworks (Sec.~\ref{sec:SC}).

\subsection{Tutorials}
\label{sec:Tut}
Some valuable tutorials for early deep learners, including books, survey papers, code tutorials, and videos, can be found at \url{http://deeplearning.net/reading-list/tutorials/}. {In addition, we list two references~\cite{net_trick1,net_trick2} which provide some general recommendations for the choice of the parameters.}

\begin{figure}[htb!]
\centering
\includegraphics[width=0.95\columnwidth]{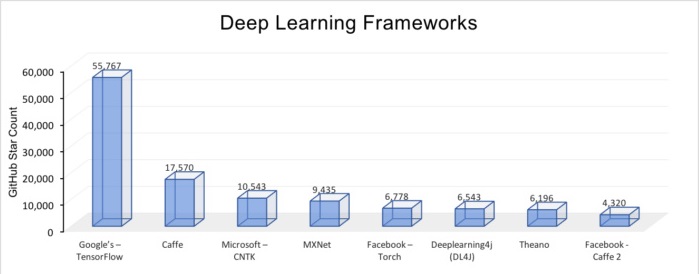}
\caption{Popular open-source deep learning frameworks. The ranking is based on the number of stars awarded by developers in GitHub. (Image source: \url{http://www.cio.com/article/3193689/artificial-intelligence/which-deep-learning-network-is-best-for-you.html})
\label{fig:deeplearning_frameworks}}
\end{figure}

\subsection{Open-source Deep Learning Frameworks}
\label{sec:ODLF}
When diving deep into deep learning, choosing an open-source framework is of great importance. Fig.~\ref{fig:deeplearning_frameworks} shows the most popular open-source deep learning frameworks, such as Caffe, Torch, Theano, TensorFlow, and Microsoft-CNTK. Since the field and surrounding technologies are relatively new and have been developing rapidly, the most common concerns amongst people who would like to work on deep learning are how these frameworks differ, where they fall short, and which ones are worth investing in. A detailed discussion of popular deep learning frameworks can be found at \url{http://www.cio.com/article/3193689/artificial-intelligence/which-deep-learning-network-is-best-for-you.html}.

\subsection{Remote Sensing Data for Training Deep Learning Models}
\label{sec:RSD}
To train deep learning methods with good generalization abilities, one needs large datasets. This is true for both fine-tuning models and training small networks from scratch, while if we consider training large architectures, one should preferably resort to pre-trained methods~\cite{castelluccio2015arxiv}.
In recent years, there is several datasets that have been made public and that can be used to train deep neural networks. Below is a non-exhaustive list.

\subsubsection{{Scene} classification (one image {is classified into} a single label)}
\begin{itemize}
\item[-] \emph{The UC Merced dataset}~\cite{Yan10}. This dataset is a collection of aerial images ($256 \times 256$ pixels in RGB space) depicting 21 land use classes. Each class comprises 100 images. Since each image comes with a single label, the dataset can be only used for image classification purposes, i.e., to classify the whole image into a single land use class. The dataset can be downloaded at \url{http://vision.ucmerced.edu/datasets/landuse.html}.

\item[-] \emph{The AID dataset}~\cite{Xia2015}. This dataset is a collection of 10,000 annotated aerial images distributed in 30 land use scene classes and can be used for image classification purpose. In comparison with the UC Merced dataset, AID contains much more images and covers a wider range of scene categories. Thus it is in line with the data requirements of modern deep learning. The dataset can be downloaded at \url{http://www.lmars.whu.edu.cn/xia/AID-project.html}.
\item[-] \emph{The NWPU-RESISC45 dataset}~\cite{cheng2017remote}. This dataset contains 31,500 aerial images spread over 45 scene classes. So far, it is the largest dataset for land use scene classification in terms of both total number of images and number of scene classes. The dataset can be obtained at \url{http://www.escience.cn/people/JunweiHan/NWPU-RESISC45.html}.
\end{itemize}

\subsubsection{{Image classification} (each pixel {of an image is classified into} a label)}
\begin{itemize}
\item[-] \emph{The Zurich Summer Dataset}~\cite{Volpi2015b}. This dataset is a collection of 20 image chips from a single large QuickBird image acquired over Zurich, Switzerland, in 2002. Each image chip is pansharpened to 0.6m resolution and 8 land use classes are presented. All images are released, along with their ground truths. The dataset can be obtained at \url{https://sites.google.com/site/michelevolpiresearch/data/zurich-dataset}.
\item[-] \emph{Zeebruges, or the Data Fusion Contest 2015 dataset}~\cite{DFCA} In 2015, the Image Analysis and Data Fusion Technical Committee of the IEEE GRSS organized a data processing competition aiming at 5-centimeter resolution land mapping. To do so, the organizers provided both a RGB aerial image and a dense ($65$ pts/m$^2$) lidar point cloud over the harbour of Zeebruges (Belgium). The data are organized on seven $10'000 \times 10'000$ pixels tiles. All the tiles have been labeled densely in 8 land classes, including land use (building, roads) and objects (vehicle, boats) classes~\cite{Lag15}. The data can be obtained from the Data and Algorithm Standard Evaluation Website (DASE) \url{http://dase.ticinumaerospace.com/}. On DASE, users can download the seven tiles and labels for five tiles. To assess models on the two remaining tiles, users can upload the classified maps on the DASE server.
\item[-] The ISPRS 2D semantic labeling challenge. The working group II/4 of the ISPRS `3D Scene Reconstruction and Analysis' provided a sub-decimeter resolution dataset over the two cities of Vaihingen and Potsdam. The data are similar to those of the Zeebruges data above, with the difference that the height information is provided as a digital surface model at the same resolution of the image data. Moreover, images are provided with an infrared channel. The dataset is also fully labeeld into six classes, including land classes (roads, meadows) and objects (cars). It also comes with a clutter class gathering all unknown objects. The Vaihingen dataset comes with 33 tiles of average size of $2000 \times 3000$ pixels. Half of the tiles come with labels. The other 17 tiles come with no labels and participants must upload classification maps for evaluation. The Potsdam dataset comes with 24 labeled tiles ($6000 \times 6000$ pixels) and 14 unlabeled ones. Both datasets can be obtained from \url{http://www2.isprs.org/commissions/comm3/wg4/semantic-labeling.html}.
\end{itemize}

\subsubsection{Registration / matching}
\begin{itemize}
\item[-] \emph{The SARptical Dataset}~\cite{Sarptical}. With the growing attention on very high resolution SAR data, the fusion of optical and SAR images in dense urban area has become an emerging and timely topic. 
Lying at the base of such fusion topic is the challenging task of the co-registration of SAR and optical images. Such two images are acquired with intrinsically different imaging geometries, and thus are nearly impossible to be co-registered without a precise 3D model of the imaged scene. SARptical is a unique dataset for SAR and optical image matching in dense urban areas. It consists of 10,000 pairs of corresponding SAR and optical image patches in central Berlin, with the center pixels of each patch pair precisely co-registered. They are generated based on co-registered 3D InSAR point clouds (which are reconstructed by SAR tomography using tens of TerraSAR-X high resolution spotlight images), and 3D optical point clouds (which are reconstructed by structure from motion followed by dense stereo matching using several UltraCam images with a ground spacing of 20cm). This dataset can be downloaded from \url{https://www.sipeo.bgu.tum.de/downloads}.
\end{itemize}

\subsection{Showcasing}
\label{sec:SC}
Starting to work with CNNs from zero might seem a titanic task. The number of models available is large and setting up an architecture from zero is challenging. In this section, we point to three showcasing example that have been recently provided by remote sensing researchers\footnote{All these examples are provided with open licenses and the corresponding papers must be acknowledged when using those codes. The rules on the respective websites apply. Please read the specific terms and conditions carefully.}. Each example uses a different deep learning library (and programming language).

\begin{itemize}
\item[-]  \emph{Deconvolution network in MatConvNet.}
The first example is released by the authors of~\cite{Vol16} and corresponds to the architecture in Fig.~\ref{fig:mitcharchi}. It exploits the MatConvNet library for MATLAB (\url{http://www.vlfeat.org/matconvnet/}). It provides a pre-trained network for both the Vaihingen and Potsdam datasets described above. The initial models are specific to remote sensing data and have been trained on each dataset separately. This example is mostly meant to show how to fine-tune an existing model in MatConvNet by training few extra iteration to improve the model weights. It can, of course, be trained from scratch by reinitializing the weights randomly. A function to test the additional images of the datasets is also provided. Overall, it allows one to reproduce the results in~\cite{Vol16} which are similar to the last column of Fig.~\ref{fig:mitchres}. By removing the deconvolutional part of the network and adding a fully connected layer at the bottleneck, one can reproduce the {\bf CNN-PC} model. If instead, one adds a spatial upsampling layer (e.g. a spatial interpolation of the bottleneck), one can also reproduce the results of the {\bf CNN-SPL} model of Fig.~\ref{fig:mitchres}. In both cases, the models must be re-trained (or at least heavily fine tuned).\newline
The code can be downloaded from \url{https://sites.google.com/site/michelevolpiresearch/codes/dense-labeling}.

\item[-] \emph{Fully convolutional (SegNet) architecture in Caff{\`e}.}
The second example is released by the authors of~\cite{Aud16} and exploits the Caff{\`e} library (\url{http://caffe.berkeleyvision.org/}). The model exploits the SegNet architecture from Kendall \emph{et al.}~\cite{Ken15}. Authors release the pre-trained model to reproduce the results of~\cite{Aud16} on the Vaihingen dataset. The network configuration, database generation and training files are given in Python.\newline
The code can be downloaded from \url{https://github.com/nshaud/DeepNetsForEO}.

\item[-] \emph{AConvNet for SAR ATR in Caffe.} The third example is released by the authors of~\cite{ChenTGRS2016}. It implements a CNN-based SAR target recognition and demonstrates via the MSTAR dataset. It includes the model configuration file and the source code for training and testing, as well as a successfully trained CNN model.\newline The code can be downloaded from \url{https://github.com/fudanxu/MSTAR-AConvNet}.

\item[-] \emph{Residual Conv-Deconv Network in TensorFlow.} The last example is released by the authors of~\cite{MouConvDeconv17,MouConvDeconvTGRS17} and shows how to build up a residual Conv-Deconv network for unsupervised spectral-spatial feature learning of hyperspectral data. It exploits TensorFlow (\url{https://www.tensorflow.org/}) and Keras (\url{https://keras.io/}) libraries. One can transfer the trained network for their own classification purpose by fine-tuning on the target data sets or obtain ``free'' object detection using the learned filters in the first residual block of the residual Conv-Deconv network.\newline
The code can be downloaded from \url{https://www.sipeo.bgu.tum.de/downloads}.

\end{itemize}

\section{Conclusion and Future Trends}\label{sec:concl}
In this paper, we reviewed the current state of the art in deep learning for remote sensing. Thanks to the enormous success encountered in several areas of research, remote sensing is also surfing the wave of deep neural networks and observing a similar trend as in other fields: deep nets are solid models that tend to improve over classical approaches using hand crafted features. {Yet, this field is still relatively young and, in the upcoming years, rapid advancement of deep learning in remote sensing is expected. Technical challenges obviously remain ahead:}

\begin{itemize}

\item {What are the further applications in remote sensing which can potentially benefit from deep learning? In general, deep nets are particularly beneficial for remote sensing problems whose physical models are complicated, e.g., nonlinear, or even not yet well understood, or/and cannot be generalized. Yet, so far, in varies remote sensing fields, most deep learning-related research has been focused on classification and detection-related tasks using a number of benchmark data sets.}

\item {Is the transferability of deep nets sufficient to extract geo-information on a global scale? Complex light scattering mechanisms in natural objects, various atmospheric scattering conditions, intra-class variability, culture-dependent features, and limited training samples make the use of deep learning for global tasks challenging~\cite{Gong2016}. To meet the need of large-scale applications, possible solutions are: never-ending learning~\cite{NeverEndingLearning}, self-taught learning~\cite{Raina07}, etc.}

\item {How to tackle problems raised by very limited annotated data in remote sensing?}
\begin{enumerate}
\item[-] {Is possible to learn deep hierarchical models for remote sensing image understanding in a weakly-supervised, semi-supervised or even unsupervised way? Here, we list a few inspiring work in machine learning and computer vision: \cite{weakly_supervised}, \cite{Johnsonicml16}, and \cite{MouConvDeconvTGRS17}.}
\item[-] {How do we benchmark the fast-growing deep-learning algorithms in remote sensing applications? Some recent initiatives include 2017 IEEE GRSS Data Fusion Contest dataset\footnote{\url{http://www.grss-ieee.org/2017-ieee-grss-data-fusion-contest/}} and Functional Map of the World Challenge dataset\footnote{\url{https://www.iarpa.gov/challenges/fmow.html}}.}
\end{enumerate}

\item {Fusion of physics-based modeling and deep neural network is a promising direction. Remote sensing imagery is a direct product of physics processes, such as light reflection, microwave scattering, etc. It has to resort to a synergy of the physics-based models which describe the a priori knowledge of the process behind imagery and newly develop artificial intelligence technologies.}

\end{itemize}
\par

{Besides focusing on technical challenges, deep learning in remote sensing opens up opportunities for new applications, such as monitoring global changes or evaluating strategies for the reduction of resources consumption, in which remote sensing can make a difference. In this context, deep learning remains an incredible toolbox that allows researcher in remote sensing to exceed the boundaries of the field, to move beyond traditional small-scale benchmarking task and tackling large-scale, real-life problems with implicit models that generalize well. The data are now here, the hardware is ready, deep learning frameworks are openly available and it is now time to design models that are tailored to big remote sensing data and their multi-modal, geo-located, multi-aspect and multi-temporal aspects that were raised in the introduction.}

\par
{On the other hand, commercial players are on the march to remote sensing and Earth observation. For example, Planet has launched about 140 small satellites which map the whole Earth daily. Standing on the paradigm shift from computational science to data-driven science, we, remote sensing experts, shall appropriately position ourselves among other data scientists, who are also trying to use deep learning for innovative remote sensing applications. This requires us, in turn and as mentioned before, to bring our domain expertises into deep learning to provide prior knowledge that is tailored to specific remote sensing problems.}
\par
{Last but not least, we advocate for efforts of the community to share data and architectures, to be able to answer the challenges of the years to come.}}

\ifCLASSOPTIONcaptionsoff
  \newpage
\fi

\bibliographystyle{IEEEtran}
\bibliography{ref_xiaoxiang_lichao,ref_feng,ref_guisong_liangpei,ref_devis,ref_friedrich}

\vskip -2\baselineskip plus -1fil

\begin{IEEEbiography}[{\includegraphics[width=1in,height=1.25in,clip,keepaspectratio]{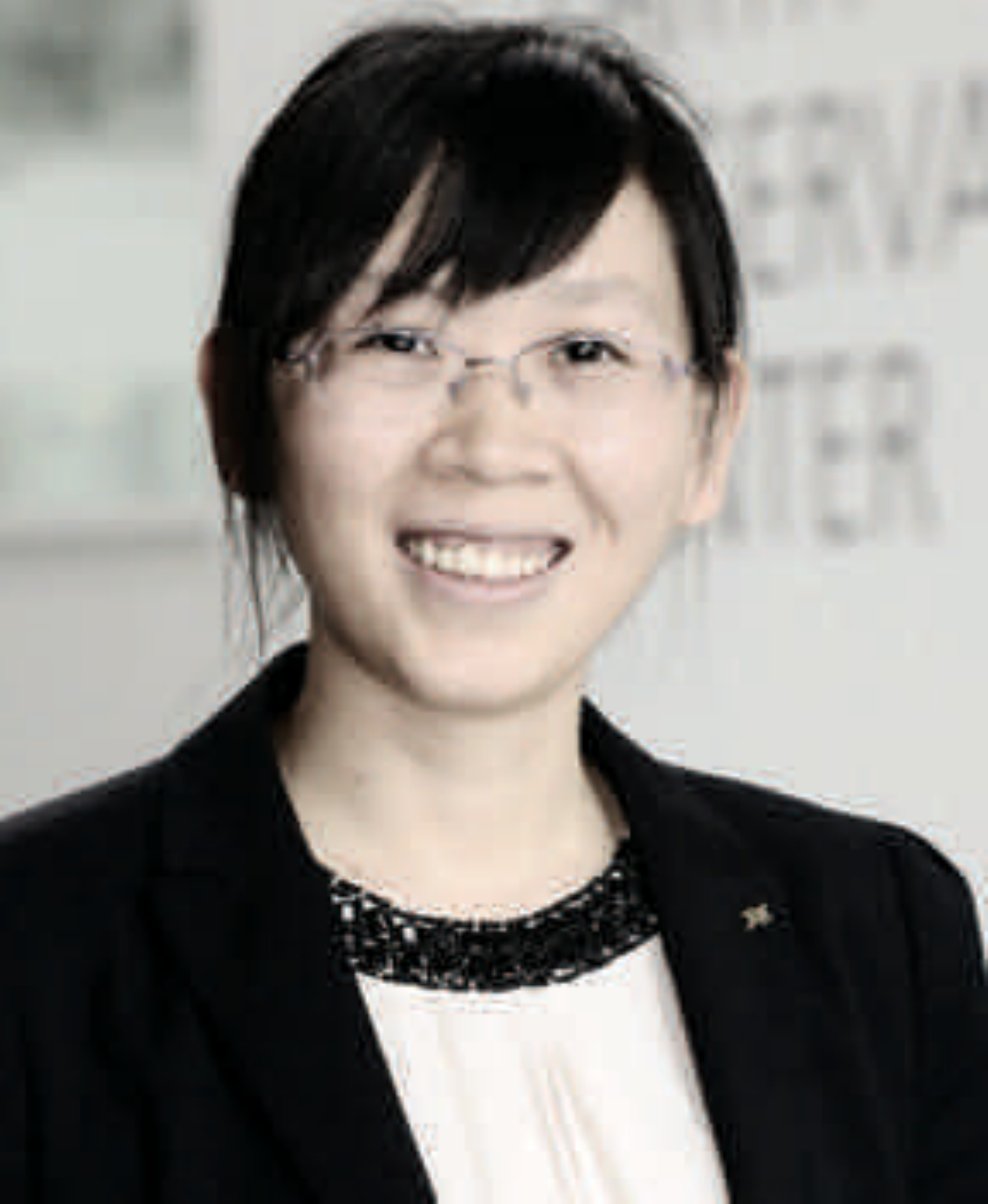}}]{Xiao Xiang Zhu}(S'10-M'12-SM'14)
received the bachelor degree in space engineering from the National University of Defense Technology (NUDT), Changsha, China, in 2006. She received the Master (M.Sc.) degree, her doctor of engineering (Dr.-Ing.) degree and her ``Habilitation'' in the field of signal processing from Technical University of Munich (TUM), Munich, Germany, in 2008, 2011 and 2013, respectively.
\par
She is currently the Professor for Signal Processing in Earth Observation (since 2015) at Technical University of Munich (TUM) and German Aerospace Center (DLR), the head of the Team Signal Analysis (since 2011) at the Remote Sensing Technology Institute, DLR, and the head of the Helmholtz Young Investigator Group "SiPEO" (since 2013), DLR and TUM. Prof. Zhu was a guest scientist or visiting professor at the Italian National Research Council (CNR-IREA), Naples, Italy, Fudan University, Shanghai, China, the University of Tokyo, Tokyo, Japan and University of California, Los Angeles, United States in 2009, 2014, 2015 and 2016, respectively. Her main research interests are: advanced InSAR techniques such as high dimensional tomographic SAR imaging and SqueeSAR; computer vision in remote sensing including object reconstruction and multi-dimensional data visualization; big data analysis in remote sensing, and modern signal processing, including innovative algorithms such as sparse reconstruction, nonlocal means filter, robust estimation and deep learning, with applications in the field of remote sensing such as multi/hyperspectral image analysis.
\par
Dr. Zhu is an associate Editor of IEEE Transactions on Geoscience and Remote Sensing.
\end{IEEEbiography}

\vskip -2\baselineskip plus -1fil

\begin{IEEEbiography}[{\includegraphics[width=1in,height=1.25in,clip,keepaspectratio]{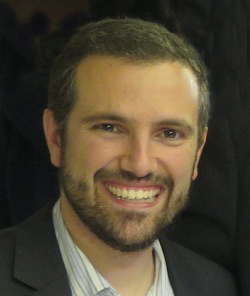}}]{Devis Tuia}
(S'07, M'09, SM'15)  received the Ph.D. from University of Lausanne in 2009.
He was a  Postdoc at the University of Val{\'e}ncia, the University of Colorado, Boulder, CO and EPFL Lausanne. Between 2014 and 2017, he was Assistant Professor at the University of Zurich. He is now Associate Professor at the GeoInformation Science and Remote Sensing Laboratory at Wageningen University, the Netherlands. He is interested in algorithms for information extraction and data fusion of geospatial data (including remote sensing) using machine learning and computer vision.
 More info on http://devis.tuia.googlepages.com/
\end{IEEEbiography}

\vskip -2\baselineskip plus -1fil

\begin{IEEEbiography}[{\includegraphics[width=1in,height=1.25in,clip,keepaspectratio]{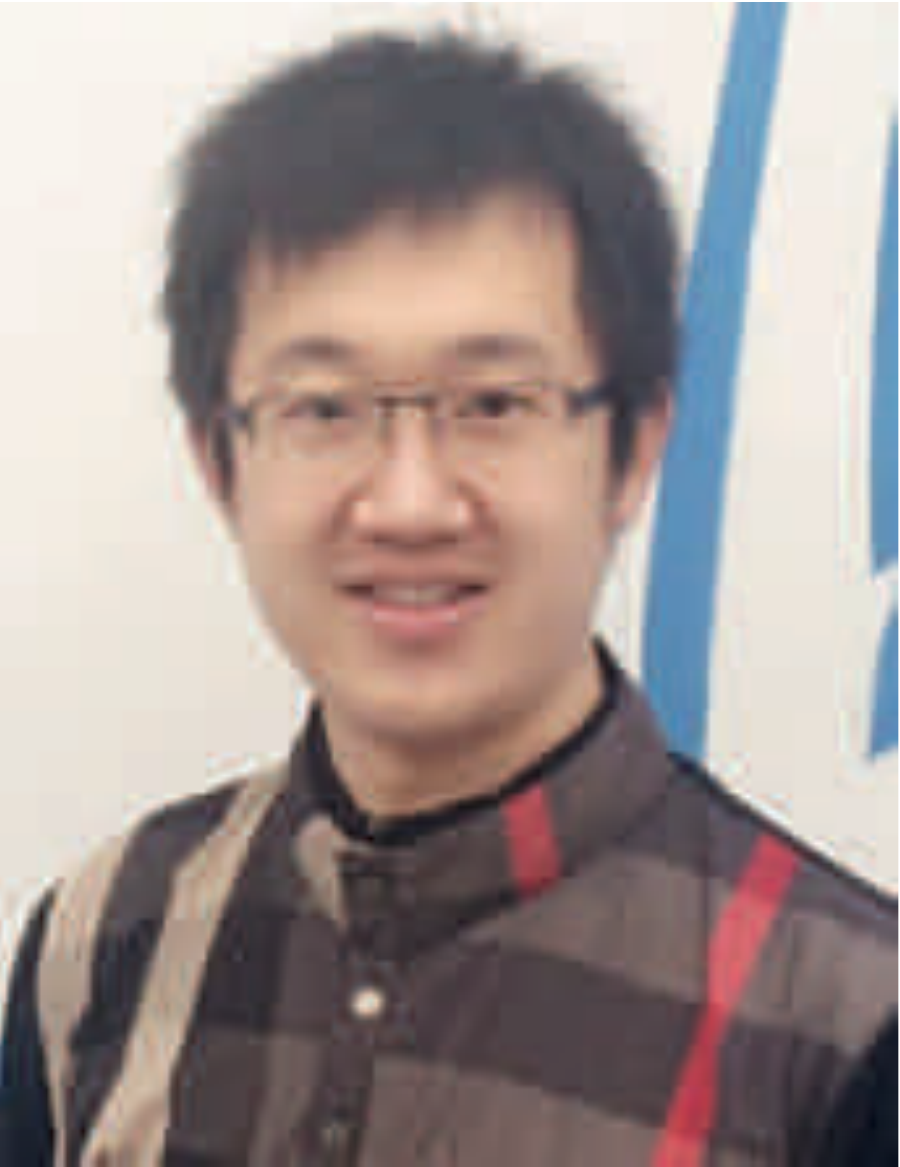}}]{Lichao Mou}(S'16)
received the Bachelor's degree in automation from the Xi'an University of Posts and Telecommunications, Xi'an, China, in 2012 and the Master's degree in signal and information processing from the University of Chinese Academy of Sciences (UCAS), China, in 2015. In 2015 he spent six months at the Computer Vision Group at the University of Freiburg in Germany. He is currently working toward the Ph.D. degree at the German Aerospace Center (DLR), Wessling, Germany, and the Technical University of Munich (TUM), Munich, Germany. His research interests include remote sensing, computer vision, and machine learning, especially remote sensing video analysis and deep networks with their applications in remote sensing.
\par
He was the recipient of the first place in the 2016 IEEE GRSS Data Fusion Contest.
\end{IEEEbiography}

\vskip -2\baselineskip plus -1fil

\begin{IEEEbiography}
[{\includegraphics[width=1in,height=1.25in,clip,keepaspectratio]{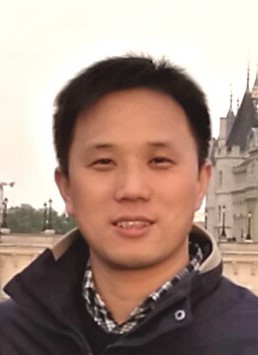}}]{Gui-Song Xia} (M'10-SM'15)
received the B.S. degree in electronic engineering and the M.S. degree in signal processing from Wuhan University,
Wuhan, China, in 2005 and 2007, respectively, and the Ph.D. degree in image processing and computer vision from the CNRS LTCI, TELECOM ParisTech, Paris, France, in 2011. Since 2011, he has been a Post-Doctoral Researcher with the Centre de Recherche en Mathmatiques de la Decision, CNRS, Paris-Dauphine University, Paris, for one and a half years.

He is currently a Professor with the State Key Laboratory of Information Engineering, Surveying, Mapping and Remote Sensing, Wuhan University, China. His current research interests include mathematical image modeling, texture synthesis, image indexing and content-based retrieval, structure from motion, perceptual grouping, and remote sensing maging.
\end{IEEEbiography}

\vskip -2\baselineskip plus -1fil

\begin{IEEEbiography}
[{\includegraphics[width=1in,height=1.25in,clip,keepaspectratio]{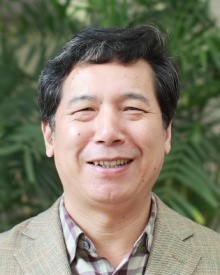}}]
{Liangpei Zhang} (M'06-SM'08)
received the B.S. degree in physics from Hunan Normal University, Changsha, China, in 1982, the M.S. degree in optics from the Xi¡¯an Institute of Optics and Precision Mechanics, Chinese Academy of Sciences, Xi¡¯an, China, in 1988, and the Ph.D. degree in photogrammetry and remote sensing from Wuhan University, Wuhan, China, in 1998.
He is currently a ¡°Chang-Jiang Scholar¡± Chair Professor at Wuhan University appointed by the Ministry of Education of China. He has published more than 500 research papers and five books. He holds 15 patents. His research interests include hyperspectral remote sensing, high-resolution remote sensing, image processing, and artificial intelligence.

Dr. Zhang is a fellow of the Institution of Engineering and Technology, an Executive Member (Board of Governors) of the China National Committee of International Geosphere-Biosphere Programme, and an Executive Member of the China Society of Image and Graphics. He was a recipient of the 2010 Best Paper Boeing Award and the 2013 Best Paper ERDAS Award from the American Society of Photogrammetry and Remote Sensing. He regularly serves as a Co-Chair of the series SPIE Conferences on Multispectral Image Processing and Pattern Recognition, Conference on Asia Remote Sensing, and many other conferences. He edits several conference proceedings, issues, and geoinformatics symposiums. He also serves as an Associate Editor of the International Journal of Ambient Computing and Intelligence, the international Journal of Image and Graphics, the International Journal of Digital Multimedia Broadcasting, the Journal of Geo-spatial Information
Science, and the Journal of Remote Sensing, and the Guest Editor of the Journal of Applied Remote Sensing and the Journal of Sensors. He is serving as an Associate Editor of the IEEE Transactions on Geoscience and Remote Sensing.
\end{IEEEbiography}

\vskip -2\baselineskip plus -1fil

\begin{IEEEbiography}[{\includegraphics[width=1in,height=1.25in,clip,keepaspectratio]{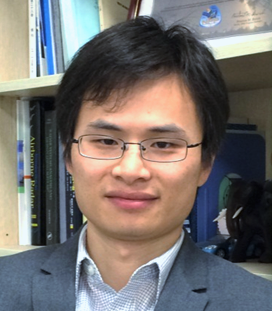}}]{Feng Xu}(S¡¯06-M¡¯08-SM¡¯14)
received the B.E. with honor in Information Engineering from Southeast University, Nanjing, China and the Ph.D. with honor in Electronic Engineering from Fudan University, Shanghai, China, in 2003 and 2008, respectively. From 2008 to 2010, he was a postdoctoral fellow with the NOAA Center for Satellite Application and Research (STAR), Camp Springs, MD. From 2010 to 2013, he was with at Intelligent Automation Inc. Rockville MD, while partly working for NASA Goddard Space Flight Center, Greenbelt, MD as a research scientist. In 2012, he was selected into China¡¯s Global Experts Recruitment Program, and subsequently returned to Fudan University in June 2013, where he currently is a professor in the school of information science and technology and the vice director of the MoE Key Lab for Information Science of Electromagnetic Waves.
\par
He has published more than 30 papers in peer-reviewed journals, co-authored 2 books, and 2 patents, among many conference papers. Among other honors, he was awarded the second-class National Nature Science Award of China in 2011. He was the 2014 recipient of the Early Career Award of IEEE Geoscience and Remote Sensing Society and the 2007 recipient of the SUMMA graduate fellowship in the advanced electromagnetics area. He currently serves as the associate editor for IEEE Geoscience and Remote Sensing Letters. He is the founding chair of IEEE GRSS Shanghai Chapter. His research interests include electromagnetic scattering theory, SAR information retrieval and radar system development.
\end{IEEEbiography}

\vskip -2\baselineskip plus -1fil

\begin{IEEEbiography}[{\includegraphics[width=1in,height=1.25in,clip,keepaspectratio]{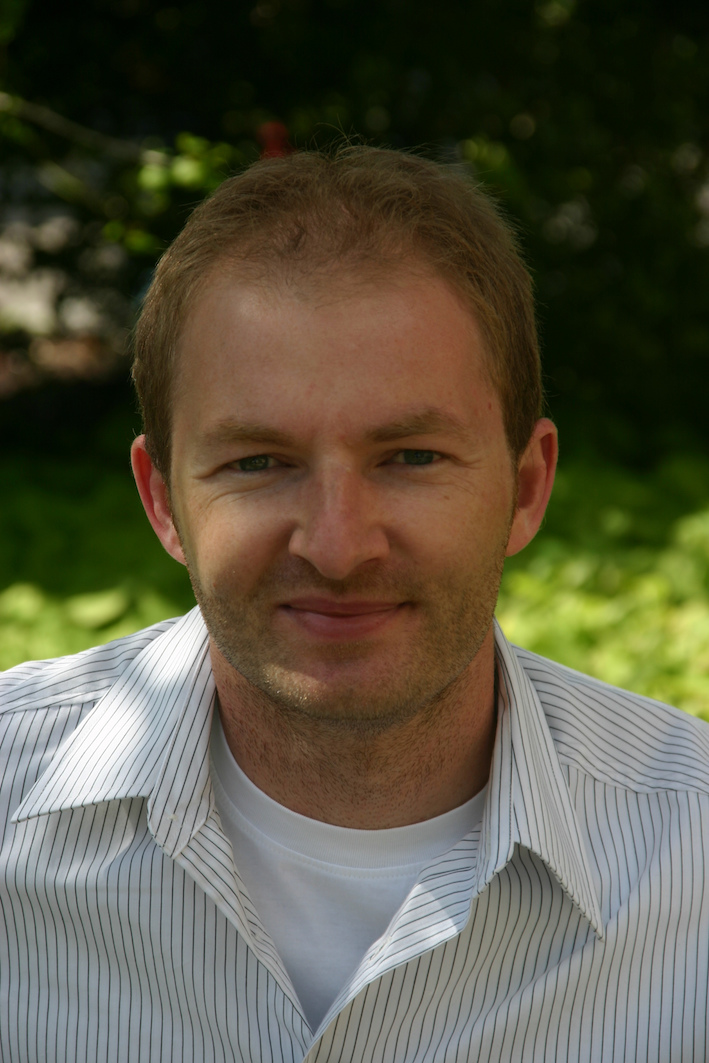}}]{Friedrich Fraundorfer}(S'10-M'12-SM'14)
received the Ph.D. degree in computer science from Graz University of Technology, Austria in 2006 working at the Institute of Computer Graphics and Vision headed by Franz Leberl and Horst Bischof.
\par
He is currently Assistant Professor at Graz University of Technology, Austria. Prior to this he had post-doc stays at the University of Kentucky (US), at the University of North Carolina at Chapel Hill (US) and at ETH Z¨¹rich (Switzerland). From 2012 to 2014 he acted as Deputy Director of the Chair of Remote Sensing Technology at the Faculty of Civil, Geo and Environmental Engineering at the Technische Universit?t M¨¹nchen.
His main research areas are 3D Computer Vision, Robot Vision, Multi View Geometry, Visual-Inertial Fusion, Micro Aerial Vehicle, Autonomous Systems, Aerial Imaging. He is the author of a well perceived two-part tutorial about visual odometry in the IEEE Robotics and Automation Magazine. His work on autonomous UAV¡¯s was awarded Best Paper Finalist at IEEE IROS 2012.
\end{IEEEbiography}

\end{document}